# A multi-strategy improved gazelle optimization algorithm for solving numerical optimization and engineering applications

**Qi Diao [1], *, Chengyue Xie [2], Yuchen Yin [3] , Hoileong Lee [4] and Haolong Yang [5]**

[1]School of Artificial Intelligence, Zhejiang Dongfang Polytechnic, Wenzhou, 325000, China
[2]Adam Smith Business School, University of Glasgow, Glasgow, Scotland, G116EY, United Kingdom
[3]Teachers College, Columbia University,525 West 120th Street, New York, NY 10027, USA
[4]Faculty of Electronic Engineering & Technology, Universiti Malaysia Perlis, 02600 Arau, Perlis, Malaysia
[5]Gina Cody School of Engineering and Computer Science, Concordia University, 1455 De Maisonneuve Blvd. W., Montreal, Quebec, Canada;
*Correspondence: diaotuanjie@163.com

**Abstract:** Aiming at the shortcomings of the gazelle optimization algorithm, such as the imbalance between exploration and exploitation and the insufficient information exchange within the population, this paper proposes a multi-strategy improved gazelle optimization algorithm (MSIGOA). To address these issues, MSIGOA proposes an iteration-based updating framework that switches between exploitation and exploration according to the optimization process, which effectively enhances the balance between local exploitation and global exploration in the optimization process and improves the convergence speed. Two adaptive parameter tuning strategies improve the applicability of the algorithm and promote a smoother optimization process. The dominant population-based restart strategy enhances the algorithm's ability to escape from local optima and avoid its premature convergence. These enhancements significantly improve the exploration and exploitation capabilities of MSIGOA, bringing superior convergence and efficiency in dealing with complex problems. In this paper, the parameter sensitivity, strategy effectiveness, convergence and stability of the proposed method are evaluated on two benchmark test sets including CEC2017 and CEC2022. Test results and statistical tests show that MSIGOA outperforms basic GOA and other advanced algorithms. On the CEC2017 and CEC2022 test sets, the proportion of functions where MSIGOA is not worse than GOA is 92.2% and 83.3%, respectively, and the proportion of functions where MSIGOA is not worse than other algorithms is 88.57% and 87.5%, respectively. Finally, the extensibility of MSIGAO is further verified by several engineering design optimization problems.

**Keywords:** Gazelle optimization algorithm; meta-heuristic algorithm; engineering optimization; adaptive; updating framework; restart strategy

## 1. Introduction

Optimization problems are widespread in all areas of the real world [1,2], and we are often faced with the challenge of making optimal decisions with limited resources and specific constraints. These challenges cut across multiple disciplines, including engineering, agronomy, medicine, economics, and management. Nowadays, it is the age of artificial intelligence, optimization methods have become a critical tool for driving technological innovation and problem solving. With the advancement of computer technology and rapid development in the field of artificial intelligence, numerous optimization methods have been innovated and revolutionized in solving scientific and engineering problems [3,4]. Traditional optimization techniques cover classical methods such as gradient - based method, integer programming, linear programming, nonlinear programming, mixed-integer programming and Newton's method [5,6]. Although these algorithms are effective in finding a globally optimal solution to a particular problem, their application is usually limited by specific conditions, such as the need for the objective function to have a convex feasible domain, continuous differentiability, or additional constraints [7]. Gradient - based methods, for instance, require the objective function to be differentiable and continuous, and they are prone to getting stuck in local optima and unable to jump out, especially in multimodal optimization landscapes [8]. Linear programming is effective for linear problems but struggles with the nonlinear relationships that are common in many real-world scenarios. However, many real-world complex optimization problems often exhibit characteristics such as nonlinearity, nonconvexity, multiple constraints, or multimodality, which make it difficult to cope with traditional methods [9]. Therefore, when dealing with large-scale and complex optimization problems, traditional optimization methods often appear to be inadequate [10].

Given the limitations of traditional optimization methods, metaheuristic algorithms have emerged to address these challenges. Such algorithms present several strengths, for example, they do not rely on derivative information, are able to exploit stochasticity to avoid falling into locally optimal solutions and exhibit higher flexibility. These properties enable metaheuristic algorithms to deal with a wider range of optimization problems, especially those with complexity, large scale and uncertainty. Typically, metaheuristic algorithms are able to find near-optimal solutions within a reasonable time frame, and even find globally optimal solutions or at least provide a near-optimal solution in some cases [11]. After extensive research and development, metaheuristic algorithms have been shown to be able to adapt to a variety of complex optimization problems and have been widely used in a variety of real-world application scenarios including, but not limited to, image segmentation [12,13], hyperparameter tuning [14–16], feature selection [17–19], health care [20,21], cluster problem [22], portfolio optimization problem [23,24], node coverage optimization for wireless sensor networks [25,26], resource allocation problem [27]and task planning problems [28–30].

Inspired by natural or social phenomena, most metaheuristic algorithms are extensively used to tackle a range of optimization problems. These optimization algorithms can be grouped into four main categories, determined by their inspiration processes, as illustrated in Figure 1.

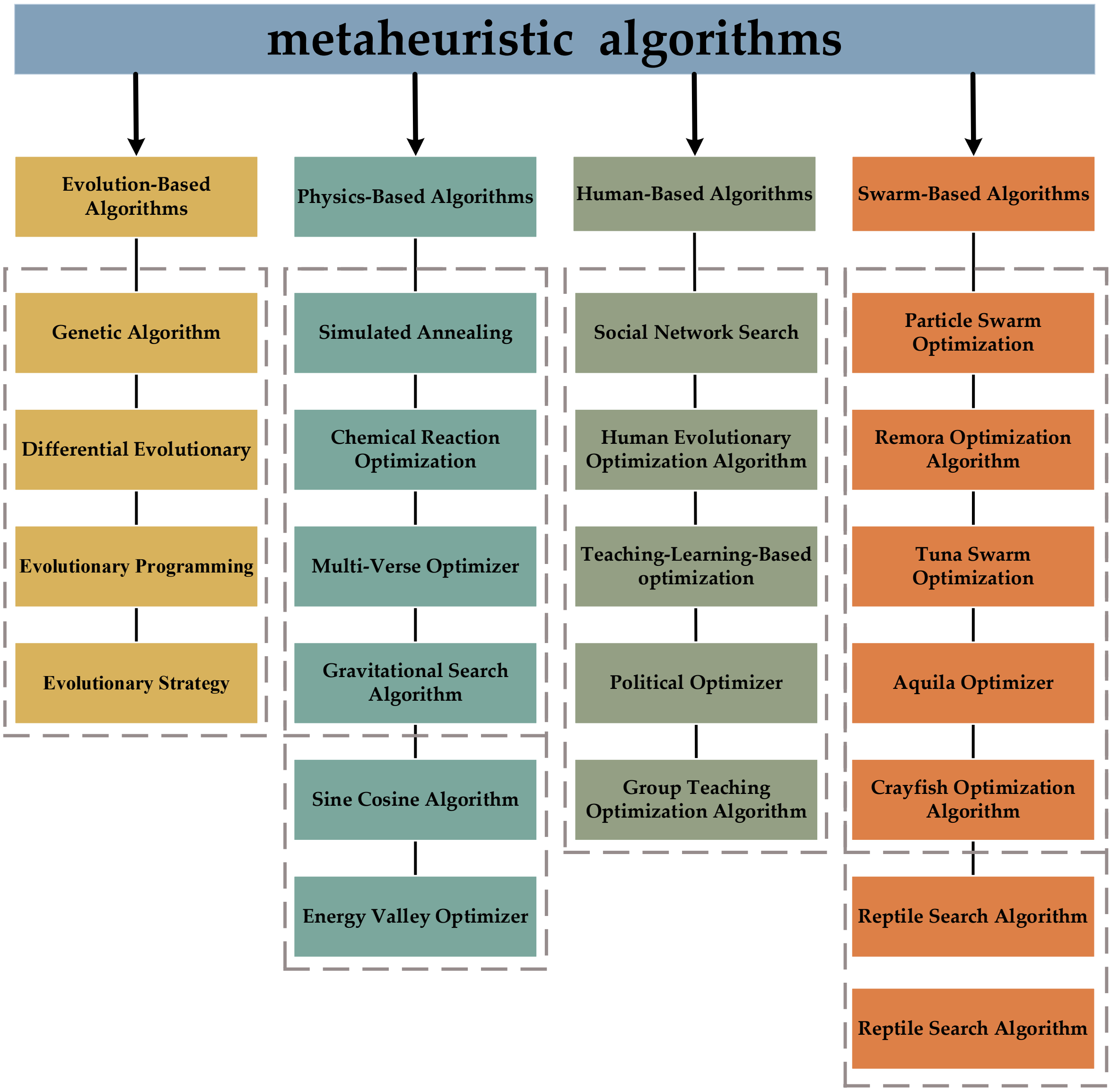

**Figure 1.** Classification of metaheuristic algorithms

(1) Evolution-Based Algorithms: These algorithms are inspired by the principles of natural evolution, including selection, mutation, and crossover. These algorithms mimic the process of natural selection, where the fittest individuals are selected to create the next generation. Examples include the Genetic Algorithm (GA) [31], Differential Evolutionary (DE) [32], Evolutionary Programming (EP) [33]and Evolutionary Strategy (ES) [34].

(2) Physics-Based Algorithms: Inspired by physical and chemical principles, examples include the Simulated Annealing (SA) [35], Chemical Reaction Optimization (CRO) [36], Multi-Verse Optimizer (MVO) [37], Gravitational Search Algorithm (GSA) [38], Sine Cosine Algorithm (SCA) [39] and Energy Valley Optimizer (EVO) [40].

(3) Human-Based Algorithms: These draw inspiration from human cooperation and behavior, such as the Social Network Search (SNS) [41], Human Evolutionary Optimization Algorithm (HEOA) [42], Teaching-Learning-Based optimization (TLBO) [43], Political Optimizer (PO) [44] and Group Teaching Optimization Algorithm (GTOA) [45].

(4) Swarm-Based Algorithms: These focus on group collaboration and information exchange to find global optima. Notable algorithms include Particle Swarm Optimization (PSO) [46], Draco Lizard Optimizer (DLO) [47], Remora Optimization Algorithm (ROA) [48], Tuna Swarm Optimization (TSO) [49], Eurasian Lynx Optimizer (ELO) [50], Aquila Optimizer (AO) [51], Crayfish Optimization Algorithm (COA) [52], Artificial Meerkat Algorithm (AMA) [53], Reptile Search Algorithm (RSA) [54] and Genghis Khan Shark Optimizer (GKSO) [55].

The Gazelle optimization algorithm (GOA) was proposed in 2022 by Agushaka et al [56] and is a swarm-based meta-heuristic algorithm. It takes inspiration from the biological behavior of gazelle herds and simulates the survival behavior of gazelles and how they escape from predators. The mathematical simulation of GOA is based on two phases (exploitation and exploration). Although GOA has solved some optimization problems such as control model parameter estimation and control [57–59], solar PV model parameter extraction [60], neural network hyper-parameter optimization [61,62], microgrid operation optimization [63], grid-connected PBES model optimization [64], and various types of battery parameter estimation [65,66], the No Free Lunch (NFL) theorem [67] suggests that this success may not generalize to other optimization problems. Although GOA has shown great potential as a metaheuristic algorithm, several limitations hinder its performance in complex optimization scenarios. A major issue is that the exploitation and exploration behaviors are randomly selected throughout the search process. This approach does not guarantee more global exploration early on and more local exploitation later on. In addition, there is a lack of information exchange between populations within the GOA, which tends to cause the algorithm to fall into a local optimum. To enhance GOA, it is essential to select an appropriate scheme that overcomes these limitations. Qin et al. [68] mixed particle swarm optimization and GOA to improve stability and enhanced population diversity using a segmented mapping initialization strategy. Mahajan et al. [69] addressed the shortcomings of GOA by integrating Levy flight strategy and random wandering strategy and solved the total harmonic distortion minimization problem. Abdollahpour et al. [70] utilized a dynamic reverse learning strategy to accelerate the GOA search process and expanded the search range by a chaotic local search strategy. Abualigah et al. [71] used an orthogonal learning approach with Rosenbrock's direct rotation strategy to improve GOA and maintain solution diversity. Maniraj et al. [64] proposed an improved GOA variant with hybrid vulture search algorithm to solve the grid-connected PV/battery storage/electric vehicle charging station optimization model. Salam et al. proposed an enhanced version of GOA. It integrates a dynamic reverse learning strategy, an adaptive inertial weight strategy, and an information exchange strategy. These improvements boost GOA's performance and resolve feature selection issues [72]. Biswas et al. integrated Differential Evolution (DE) into the Gazelle Optimization Algorithm (GOA) to combine the strengths of both algorithms. This hybrid approach enhances the global and local search capabilities for complex optimization problems [73]. The existing literature indicates that the GOA algorithm faces several challenges that affect its performance. These include a tendency to fall into local optima, slow convergence speed, and loss of population diversity. Additionally, while robust search methods and hybrid techniques can enhance GOA's performance, issues such as easy trapping in local optima, premature convergence, and poor exploitation ability still persist. Furthermore, the No Free Lunch (NFL) theorem suggests that while an algorithm may be effective for certain optimization problems, it may not be suitable for others due to their unique characteristics.

In response to the limitations identified in the aforementioned GOA, this paper proposes a multi-strategy improved gazelle optimization algorithm (MSIGOA). We propose a new iteration-based search framework in MSIGOA. This framework can assist GOA to explore in the early stage, combine both global search and local development in the middle stage, and focus on local search in the end stage. This framework well coordinates the switch between exploration and exploitation. In addition, two adaptive parameter tuning strategies help GOA achieve a harmonious balance between exploration and exploitation. Finally, we propose a restart strategy based on dominant populations to help GOA to escape from local optima. In order to comprehensively evaluate the performance of MSIGOA, we conducted extensive tests using 29 CEC2017 benchmark test functions, 12 CEC2022 benchmark test functions and 3 engineering design optimization problems. The primary contributions of this paper are outlined as follows:

(1) Iteration-based update framework promotes a better balance between development and exploration and provides a smoother optimization process.

(2) Adaptive parameter tuning strategy adjusts parameters at different stages to enhance the adaptability of the algorithm.

(3) Advantageous population-based restart strategy enhances population diversity, reduces the risk of falling into local optimality and improves the quality of the solution.

(4) Adequate and comprehensive experiments: 8 comparison algorithms, 2 benchmark test sets and 3 engineering design optimization problems.

(5) Significant advantages: On the CEC2017 and CEC2022 test sets, the proportion of functions where MSIGOA is not worse than GOA is 92.2% and 83.3%, respectively, and the proportion of functions where MSIGOA is not worse than other algorithms is 88.57% and 87.5%, respectively.

The paper is organized as follows: Section 2 delves into the architecture and details of the GOA algorithm. Section 3 proposes three mechanisms to enhance the GOA algorithm: iteration-based updating framework, adaptive parameter tuning strategy, and dominant population-based restart mechanism. Section 4 demonstrates the parameter sensitivity and ablation experiments of MSIGOA, a comprehensive evaluation of the proposed method on the CEC2017 test set, the CEC2022 test set and the engineering design optimization problem, and a statistical analysis of the experimental results. Finally, Section 5 reviews the findings of the paper and provides an outlook for future research.

## 2. An overview of gazelle optimization algorithm (GOA)

The Gazelle Optimization Algorithm (GOA), introduced by Agushaka et al. in 2022, emulates the evasion tactics of gazelles against predators, presenting a high-efficiency methodology. The developmental stages of this algorithm simulate a gazelle peacefully grazing on flat ground in the absence of predators or when predators are not actively tracking it. Nevertheless, upon predator detection, GOA shifts into an exploration mode. This mode includes the gazelle's escape maneuvers and strategic outmaneuvering of the predator to secure a safe haven. These dual stages are iterated in accordance with predefined termination conditions to uncover the optimal solution for the optimization challenge. The GOA consists of three phases: global search, local search, and gazelle escape.

### *2.1. Initialization Process*

GOA is a swarm-based optimization algorithm that, similar to most algorithms, initializes the search agents by using the number of populations within a specified boundary of the solution space. The search agents form a matrix with $N$ rows and $D$ columns, where $N$ is a quantity of candidate solutions and $D$ denotes the dimensionality of the search the problem spaces , as shown in Equation 1. In GOA, the range of candidate solution values is determined by the constraints on the upper ($Ub$) and lower ($Lb$) limits of the problem. The matrix $X$ consists of the positions of the candidate population, where each search agent is randomly generated using Equation 2. Specifically, $X_i$ denotes the position of the $i^{th}$ search agent. Assuming the objective function is $F$ , the objective function value of the $i^{th}$ search agent can be calculated as below.

$$X = [X_1, X_2, \cdots, X_N] = \begin{bmatrix} X_{1,1} \cdots X_{1,j} \cdots X_{1,D} \\ \vdots \cdots \vdots \cdots \vdots \\ X_{i,1} \cdots X_{i,j} \cdots X_{i,D} \\ \vdots \cdots \vdots \cdots \vdots \\ X_{N,1} \cdots X_{N,j} \cdots X_{N,D} \end{bmatrix} \tag{1}$$

$$X_i = lb + rand(1, D) \times (Ub - Lb) \tag{2}$$

$$Fitness_i = F(X_i) \tag{3}$$

*2.2. Exploitation Process*

During the exploitation phase, it is assumed that the gazelle is in a free foraging state, with no predators emerging or predators stalking the gazelle. During this period, the movement of the gazelle follows the principle of Brownian motion. Brownian motion is characterized by uniform and controlled steps being used to efficiently cover the neighborhood of the domain. The mathematical model of this behavior is shown in Equation 4.

$$X_i^{t+1} = X_i^t + s \times rand(1, D) \times R_B \times (Elite - R_B \times X_i^t) \tag{4}$$

In this equation, $X_i^t$ represents the position of the $i^{th}$ individual in iteration $t$ , and $X_i^{t+1}$ represents its updated position. $s$ is the speed of gazelles in the period of grazing with a value of 0.88. $R_B$ is a vector that is formulated of random numbers of Brownian motion. $Elite$ denotes the candidate solution that has performed optimally throughout the entire search process. $t$ is the current iteration.

*2.3. Exploration Process*

When a gazelle encounters a predator, the exploration phase occurs. This phase primarily mimics the escape behavior of gazelles. Escape behavior is divided into two stages. In each escape, gazelles choose one of two opposite directions. This choice is determined by the parity of the number of iterations. In GOA, we assume that gazelles choose the Levy flight mechanism to escape in the early stages of predator detection. Over time, the gazelle transitions to adopting Brownian motion, which simulates irregular movements when avoiding a predator.

The gazelle adopts Lévy flight before spotting any predators, as shown in Equation 5.

$$X_i^{t+1} = X_i^t + s \times \mu \times rand(1, D) \times R_L \times (Elite - R_L \times X_i^t) \tag{5}$$

$$R_L = 0.05 \times \frac{z}{|y|^{\frac{1}{\alpha}}} \tag{6}$$

In this equation, $\mu$ is either 1 or -1, depending on the parity of the number of iterations, that is, $\mu$ is 1 when the number of iterations is odd-number and vice versa for -1. $R_L$ is a vector that is formulated of random numbers of Lévy distribution. The expression for $R_L$ is given by Equation 6, where $z = normal(0, \sigma_z^2)$ , $y = normal(0, \sigma_y^2)$ , $\sigma_z = \frac{\Gamma(1+\alpha) \times \sin\left(\frac{\pi\alpha}{2}\right)}{\Gamma\left(\frac{1+\alpha}{2}\right) \times \eta \times 2^{\left(\frac{\alpha-1}{2}\right)}}$ , $\sigma_y = 1$ , $\alpha = 1.5$. $\Gamma(\cdot)$ denotes the gamma function.

The gazelle adopts Brownian motion after spotting any predators, as shown in Equation 7.

$$X_i^{t+1} = Elite + s \times \mu \times CF \times R_B \times (R_L \times Elite - X_i^t) \tag{7}$$

$$CF = \left(\frac{t}{T}\right)^{\frac{2t}{T}} \tag{8}$$

In this equation, $T$ is the maximum iteration.

*2.4. Escape Process*

The studies show that the predators are able to catch the gazelles specially the Mongolian gazelles only 34% most of the time or the gazelles can survive at a rate of 0.66 [74]. PSR (Predator Success Rate) is the probability of predators successfully capturing gazelles that can be formulated by the following Equation 9.

$$X_i^{t+1} = \begin{cases} X_i^t + CF \times U \times \left(Lb + rand(1,D) \times \left(Ub - Lb\right)\right), r_2 \le PSR \\ X_i^t + \left(PSR \times \left(1 - r_1\right) + r_1\right)\left(X_A^t - X_B^t\right), r_2 > PSR \end{cases} \tag{9}$$

In this equation, $U$ is a binary vector which obey $U_j = \begin{cases} 0, rand < 0.34 \\ 1, rand \ge 0.34 \end{cases}$,is constructed by generating a random number in the range [0,1]. $X_A^t$ and $X_B^t$ are two random selected agents. $r_1$ and $r_2$ are random numbers in the range 0 to 1.

## 3. A framework of newly proposed MSIGOA

As can be seen in Section 3, GOA randomly switches between exploitation and exploration behaviors during the optimization process, making it difficult to reconcile the balance between global and local search. In addition, Levy flight and Brownian motion are not adjusted with the search process, making it difficult for the GOA to show favorable adaptability. The lack of sufficient information exchange among the GOA population suffers from problems such as easy to fall into local optimum. To address these shortcomings, three improvement methods are proposed in this paper, including iteration-based updating framework, adaptive parameter tuning strategy and dominant population-based restart mechanism. These improvement strategies aim to enhance the algorithm's global search capability, suitability, convergence speed and accuracy, thus enhancing the performance of GOA in solving optimization problems.

*3.1. Iteration-based updating framework (IBUF)*

GOA randomly selects either exploitation or exploration at each iteration. However, the meta-heuristic algorithm wants to explore a wide range of promising regions earlier in the process and concentrate more on small regions for precise exploitation in the later process. This search mechanism of GOA makes it difficult to balance exploitation and exploration. For this reason, this paper proposes an iteration-based update framework. This framework can help GOA to conduct more global search in the early stage and more local exploitation in the later stage, while it can have both exploitation and exploration capabilities in the middle stage. Specifically, IUBF divides the optimization process into three segments. The first stage mainly performs global search, which helps the algorithm to explore the problem area more in the early stage and enhance the population diversity. The second stage performs both global and local search, which ensures smoother switching of the search mechanism. In the third phase the algorithm will concentrate on local exploitation, which improves the convergence accuracy of the algorithm. For the characteristics of the three stages, Equation 4, Equation 6 and Equation 7 are adapted and applied to the corresponding stages as follows.

In IBUF, the first phase is defined as the first one-third of the iteration period, i.e., when $t < \frac{T}{3}$, GOA performs the global search strategy according to Equation 4. The second phase is defined as the middle third of the iteration period, i.e., when $\frac{T}{3} < t \le \frac{2T}{3}$, the GOA performs the global search strategy according to Equation 5 and Equation 7. The third phase is defined as the second third of the iteration period, i.e., when $t \ge \frac{2T}{3}$, the GOA performs a localized search strategy according to Equation 5 and Equation 10.

$$X_i^{t+1} = Elite + s \times \mu \times CF \times R_L \times \left(R_L \times Elite - X_i^t\right) \tag{10}$$

*3.2. Adaptive parameter tuning strategy (APTS)*

Brownian motion can cover the problem area with more uniform and controlled step sizes. However, this is detrimental to population search and lacks flexibility in the middle and late periods [10]. The Lévy flight strategy searches the space mainly with small step sizes associated with long jumps and is able to explore other regions of the domain due to the larger step sizes, but it is not able to cover all the regions of the domain individually, and is deficient in balancing exploitation and exploration. Therefore, we propose two adaptive parameter tuning strategies. For Brownian motion, we introduce a tuning factor $\left(1-\frac{t}{T}\right)^{\frac{t}{T}}$ to enhance its flexibility and allow it to explore the search space more flexibly. In addition, the algorithm is able to adaptively adjust the parameters at different stages to improve the global search capability of the algorithm and prevent the algorithm from falling into a local optimum. This enhancement increases the adaptability of the algorithm. For the Levy flight, we introduce an adjustment factor $\left(1-\frac{t}{T}\right)$ to dynamically adjust the parameters of the Levy flight. This introduction can be considered as an adaptive step-size adjustment mechanism that helps to achieve a better balance between exploration and exploitation. It helps to avoid drastic changes in step size in later iterations and ensures a smoother optimization process. The improved Brownian motion and Levy flight are shown below.

$$R_B = R_B \times \left(1-\frac{t}{T}\right)^{\frac{t}{T}} \tag{11}$$

$$R_L = R_L \times \left(1-\frac{t}{T}\right) \tag{12}$$

*3.3. Dominant population-based restart mechanism (DPRM)*

The problem centers on the fact that GOA lacks information exchange between populations. Most of the search agents establish connections with the optimal individuals to develop and explore around them, which can speed up the convergence. However, blindly following the optimal individual weakens the population diversity of the algorithm and increases the risk of falling into a local optimum[75]. In order to improve the quality of candidate agents and tap the guidance of the dominant population, this paper proposes a dominant population-based restart mechanism (DPRM). The mechanism contains three reference points: the optimal agent position, the weighted position of the dominant population, and the agent itself. The addition of the optimal agent can accelerate the convergence of the algorithm. The addition of the dominant group promotes the movement of the agents to promising regions. The inclusion of the agent itself enhances the population diversity. The integration of the three reference points fully utilizes the effective information of the population and significantly improves the performance of the GOA. The DPRM can be calculated from Equation 13.

$$X_i = \frac{\left(X_i + X_d + Elite\right)}{3} + G_i, G_i \sim N\left(0, C\right) \tag{13}$$

$$X_d = \frac{1}{N_d} \times \sum_{i=1}^{N_d} \left( \frac{\ln(N_d + 0.5) - \ln(i)}{\sum_{i=1}^{N_d} (\ln(N_d + 0.5) - \ln(i))} \times X_i \right), X_i \in P_d \tag{14}$$

$$G = \frac{1}{N_d} \times \sum_{i=1}^{N_d} \left(X_i - X_d\right) \times \left(X_i - X_d\right)^T, X_i \in P_d \tag{15}$$

In these equations, $X_d$ is the weighted average position of the dominant population $P_d$. $N_d$ denotes the number of dominant populations. Historical information about the population can likewise facilitate the algorithm's search toward promising areas. The dominant population $P_d$ consists of the top fifty percent of agents that perform better after each iteration. Consider that the effectiveness of historical information diminishes as iterations progress. Therefore, the first-in-first-out principle is employed to scale the total number of dominant groups. For the setting of $N_d$, it will be discussed in the experimental section.

*3.4. Implementation steps for MSIGOA*

By integrating the above three strategies into GOA, MSIGOA is proposed. MSIGOA is divided into the following four phases. (1) Initialization phase. (2) Exploitation and exploration phase. In this phase, RB and RL are adjusted with the number of iterations according to the adaptive parameters tuning strategy. The population selects the corresponding search method according to the number of iterations under the iteration-based updating framework. (3) Escape phase. (4) Newly addition restart phase. The dominant population-based restart mechanism is utilized to enrich the population diversity and improve the GOA performance. The flowchart of MSIGOA is given in Figure 2. Algorithm 1 shows the pseudo-code of MSIGOA.

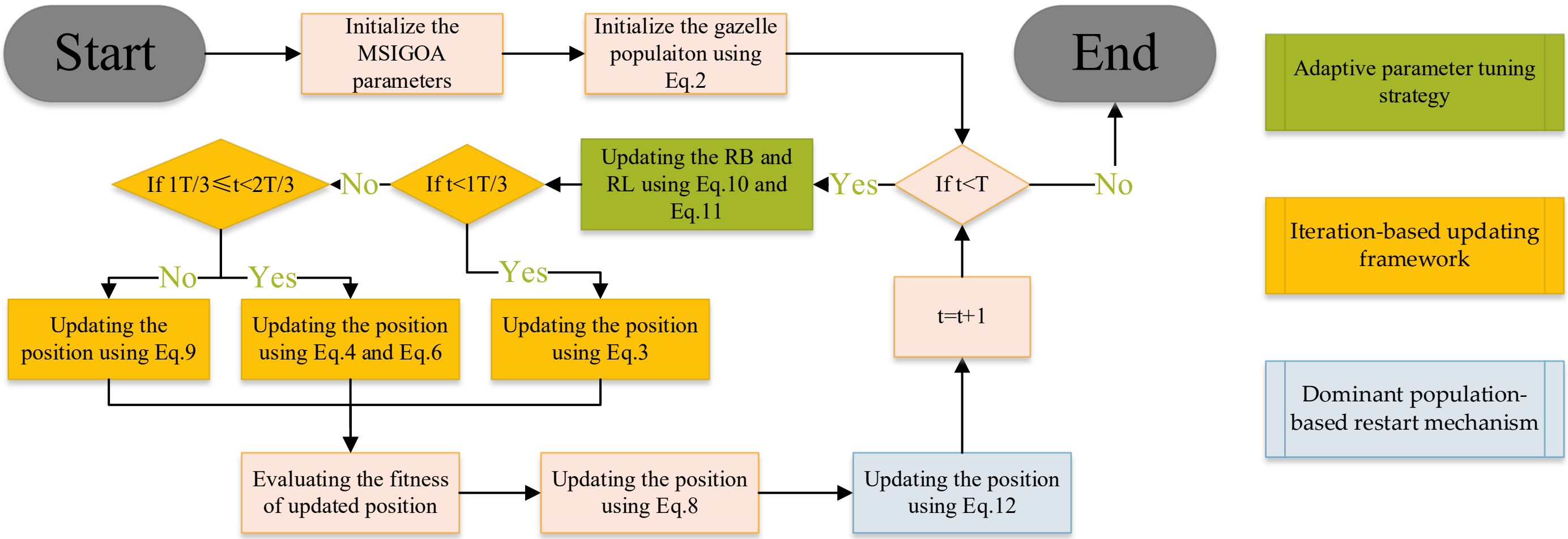

**Figure 2.** Flow chart of MSIGOA

| **Algorithm 1 Multi-strategy improved gazelle optimization algorithm(MSIGOA)** |
|---|
| Initialization the parameters： $t = 0$, $T$, $N$, $N_d$, $PSR$, |
| Generate an initial population randomly according to Eq.2 |
| While ($t$<$T$) do |
| For $i$=1:$N$ |
| If $t < T/3$ |
| Updating gazelles position based on Eq.4 |
| Else if $T/3 \le t < 2T/3$ |
| If $i < N/2$ |
| Update gazelles position based on Eq.5 |
| Else |
| Update gazelles position based on Eq.7 |
| End if |
| Else |
| Update gazelles position based on Eq.10 |
| End if |
| Update gazelles position based on Eq.9 |
| Applying DPRM and update gazelles position based on Eq.13 |
| End for |
| $t$=$t$+1 |
| End while |

*3.5. Analysis of MSIGOA complexity*

The time complexity of MSIGOA is evaluated with Big-O notation. First, assuming the population size is $N$, a maximum iteration count of $T$, and an optimization problem dimension of $D$, the time complexity of GOA is $O(2N \times T \times D)$. Next, the complexity of MSIGOA is analyzed:

The iteration-based updating framework ensures that each individual chooses only one search method at each iteration and hence does not increase the time complexity. The adaptive parameter tuning strategy only changes the values of the parameters and does not increase the time complexity. Assuming that the restart mechanism is used by $N_a$ agents per iteration, the time complexity of the dominant population-based restart mechanism is $O(T \times N_a \times D)$. In summary, the overall time complexity of the MSIGOA is $O((2N + N_a) \times T \times D)$. Therefore, the time complexity of MSIGOA is slightly increased compared to GOA.

## 4. A framework of newly proposed MSIGOA

This section validates the performance of the proposed MSIGOA through several experiments. Section 4.1 describes the parameter settings of the benchmark test functions and comparison algorithms employed. Section 4.2 discusses the parameter sensitivity of the dominant population-based restart mechanism implemented in MSIGOA. In Section 4.3, we perform ablation experiments to evaluate the effectiveness of each improvement strategy. Section 4.4 presents the results of the comparison between MSIGOA and the competitors on the CEC2017 and CEC2022 benchmarking functions. Section 4.5 shows the performance of MSIGOA on several engineering design optimization problems.

### *4.1. Experiment setting*

The proposed MSIGOA algorithm has been implemented in MATLAB R2022a on a computer system equipped with an AMD R9 7950X processor running at 4.50 GHz and 32 GB of RAM. Typically, an optimization algorithm performs well on certain test functions but poorly on others. Thus, a diverse set of test functions is required to assess the overall effectiveness of the improved algorithm. In this study, a comprehensive set of 41 test functions from the CEC2017 and CEC2022 is selected. These functions comprise two set of 41 test functions encompassing various characteristics such as unimodal, multimodal, hybrid, and composition functions. Table 1 and Table 2 give detailed descriptions of these benchmark functions. The unimodal functions focus on single-peak landscapes, which are simpler but essential for testing the algorithm's ability to converge to a global optimum. The multimodal functions include multiple peaks and valleys, making them suitable for evaluating the algorithm's exploration and exploitation balance. The hybrid and composition functions have complex structures that fully check the comprehensive performance of an algorithm.

**Table 1.** CEC 2017 benchmark test functions.

| Type | No. | Range value |
|---|---|---|
| Unimodal | F1-F2 | [-100,100] |
| Multimodal | F3-F9 | [-100,100] |
| Hybrid | F10-F19 | [-100,100] |
| Composition | F20-F29 | [-100,100] |

**Table 2.** CEC benchmark test functions.

| Type | No. | Range value |
|---|---|---|
| Unimodal | F1 | [-100,100] |
| Basic | F2-F5 | [-100,100] |
| Hybrid | F6-F8 | [-100,100] |
| Composition | F9-F12 | [-100,100] |

For comparison, nine algorithms are selected include basic GOA, MRFO[76], FTTA[77], NMPA[78], EOSMA[79], BEESO[80], FDBARO[81], IDE-EDA[82], APSM-jSO[83]. The parameters for each algorithm are established following their original settings, as detailed in Table 3. Furthermore, the number of iterations for all algorithms is set to 500, the initial population size is set to 30, and each algorithm is run 51 times, and two evaluation metrics were utilized to compare and analyze the optimization performance of each method intuitively: average value (Mean) and standard deviation (Std).

**Table 3.** Parameter settings of each algorithm.

| Algorithm | Parameter value |
|---|---|
| MSIGOA | $s = 0.88, PSR = 0.34, N_d = 25D$ |
| GOA | $s = 0.88, PSR = 0.34$ |
| MRFO | $S = 2$ |
| FTTA | $stu = 0.3, com = 0.3, error = 0.001, tn = 3, zb = 4$ |
| NMPA | $FADs = 0.2, P = 0.5$ |
| EOSMA | $V = 1, a_1 = 2, GP = 0.5, z = 0.6, q = 0.2$ |
| BEESO | $Tr_1 = 0.25, Tr_2 = 0.6, C_1 = 0.5, C_2 = 0.05, C_3 = 2$ |
| FDBARO | No parameter |
| IDE-EDA | $k = 3, F = 0.3, CR = 0.8, iChy = 1$ |
| APSM-jSO | $F = 0.3, cr = 0.8, h = 6, Ar = 1.3, iM = 1$ |

### *4.2. Parameter sensitivity analysis on* $N_d$

The parameter $N_d$ is an important parameter for the dominant population-based restart mechanism. It determines the number of dominant populations used by the DPRM and how much historical information it contains, affecting the population evolutionary direction. In order to fully utilize the performance of MSIGOA, we discuss the effects of six different $N_d$ ($N_d = 1D, 5D, 10D, 15D, 20D, 25D$) on MSIGOA. The experiments were performed on the CEC2017 test set. Table 4 demonstrates the statistical results based on the Friedman test. Figure 3 presents the results of the Wilcoxon rank sum test between MSIGOA with different $N_d$ values and GOA.

**Table 4.** The Friedman ranking of different parameter Nd

| | GOA | $N_d$=1×$D$ | $N_d$=5×$D$ | $N_d$=10×$D$ | $N_d$=15×$D$ | $N_d$=20×$D$ | $N_d$=25×$D$ |
|---|---|---|---|---|---|---|---|
| D=10 | 3.90 | 6.10 | 4.86 | 4.69 | 3.62 | 2.79 | **2.03** |
| D=30 | 5.28 | 6.28 | 5.45 | 3.93 | 2.90 | 2.21 | **1.97** |
| D=50 | 5.93 | 6.17 | 4.41 | 3.41 | 3.03 | 2.55 | **2.48** |
| D=100 | 5.69 | 5.72 | 3.55 | **2.93** | 2.97 | 3.17 | 3.97 |
| Mean rank | 5.20 | 6.07 | 4.57 | 3.74 | 3.13 | 2.68 | **2.61** |

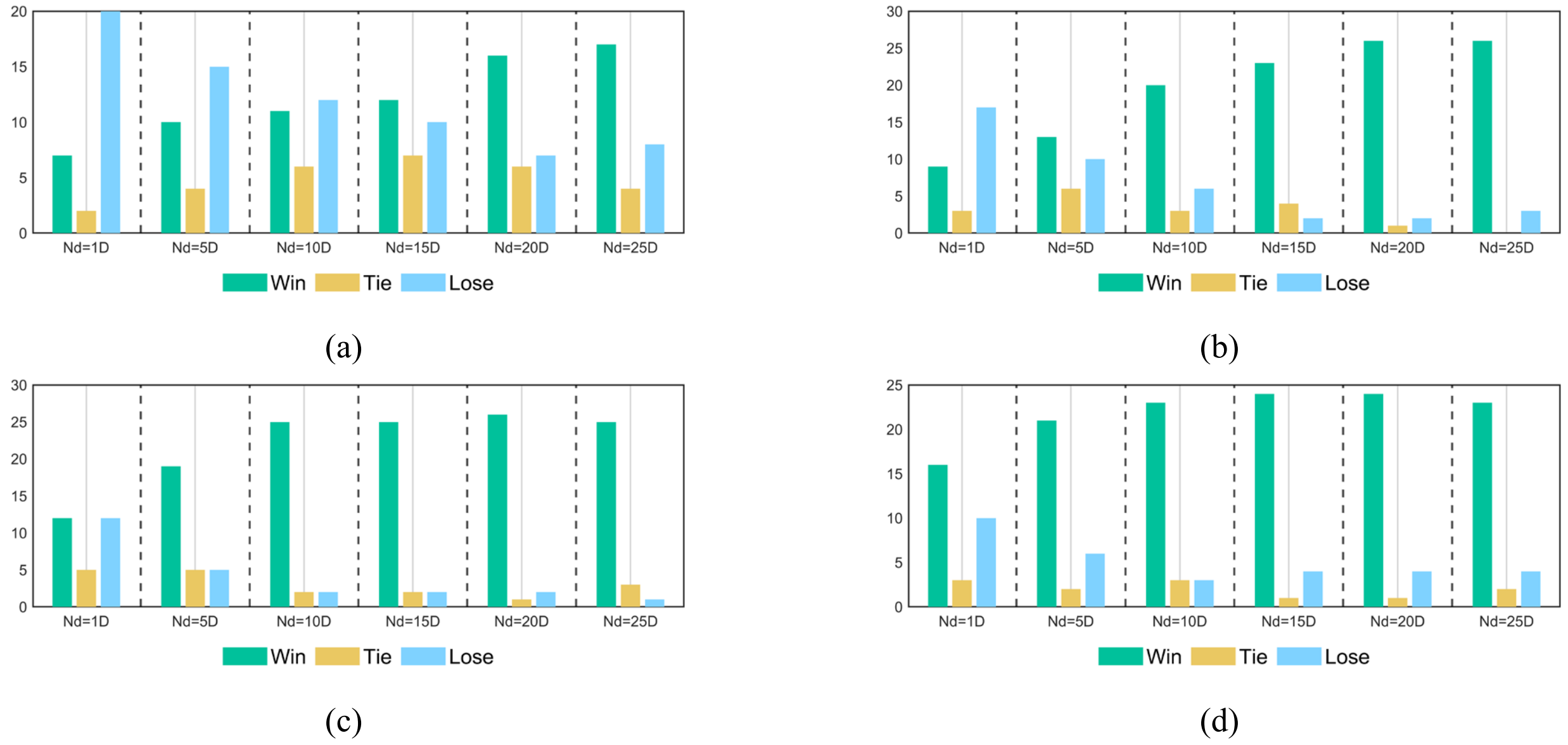

**Figure 3.** Wilcoxon rank sum test results for the parameters $N_d$. (a)CEC2017 10D, (b)CEC2017 30D, (c)CEC2017 50D, (d)CEC2017 100D.

In Table 4, the optimal values for each case are shown in bold. For CEC2017 10D, the DPRM is not as effective when the number of dominant groups is small ($N_d = 1D, 5D, 10D,$), and it shows satisfactory performance when takes a large value ($N_d = 15D, 20D, 25D$). For CEC2017 30D, DPRM fails to improve the performance of GOA when $N_d$ is *10D* and *30D*, and when $N_d$ is further increased, DPRM helps GOA to get high quality solutions. For CEC2017 50D and 100D, DPRM is ineffective only when $N_d$=10D, and all other quantities can enhance the GOA. According to Figure 3, we further find that even for $N_d$=10D, DPRM still improves the performance of GOA when the dimension is large enough. For CEC2017 50D, the number of functions that MSIGOA outperforms the other at $N_d$=1D and GOA is the same. For CEC2017 50D, the number of functions in which MSIGOA outperforms GOA at $N_d$=1D is even more than the number of functions in which GOA outperforms MSIGOA.

The above results suggest that DPRM needs a sufficiently large dominant population to full $N_d$ y utilize the mechanism. Since the Nd setting is associated with the problem dimension, the number of dominant groups in low dimensions will be insufficient as a result, which leads to the fact that multiple Nd values in low dimensions are unable to enhance the performance of the GOA. $N_d$=25D achieves the first place in all three cases, and $N_d$=10D achieves the best ranking in 100D, but the overall performance of $N_d$=25D is much better. Therefore, in this paper, $N_d$ was set to 25D for subsequent experiments.

*4.3. Ablation experiment analysis on each strategy*

In this section, we evaluate the impact of each improvement method on GOA. In our experiments, we used two sets of functions that are widely used for algorithm performance comparison and testing, including the CEC2017 and CEC2022 benchmark test sets, to perform the ablation experiments. Table 5 describes the details of the six variants. The first column of the table lists the names of the variants, and the first row describes the strategy used for each variant. “Y” denotes a variant that utilizes the mechanism, while ‘N’ denotes no mechanism. Specifically, GOA-1 contains only an iteration-based update framework.GOA-2 uses only an adaptive parameter tuning strategy.GOA-3 integrates a restart mechanism based on dominant populations. Compared to MSIGOA, GOA-12 removes the restart mechanism, GOA-13 removes the adaptive strategy, and GOA-23 removes the update framework. By comparing with these variants, we can verify whether each improvement strategy has a positive impact on the performance of MSIGOA.

**Table 5.** Various GOA variants with three strategies

| Algorithm | IBUF | APTS | DPRM |
|---|---|---|---|
| GOA-1 | Y | N | N |
| GOA-2 | N | Y | N |
| GOA-3 | N | N | Y |
| GOA-12 | Y | Y | N |
| GOA-13 | Y | N | Y |
| GOA-23 | N | Y | Y |
| MSIGOA | Y | Y | Y |

Table 6 and Figure 4 summarize the data from the ablation experiments based on Friedman test and Wilcoxon rank sum test, respectively. According to the p-value of Friedman's test in the last column in Table 6, there is a significant difference between all GOA variants and GOA. Based on Figure 4, the total number of “+” for each variant is more than the number of “-” in each case, which indicates that the GOA variants are superior to GOA in terms of overall performance. Specifically, the MSIGOA that integrates all three strategies performs the best under all dimensions. The statistical results obtained are analyzed in detail below.

The effect of IBUF on GOA can be evaluated by analyzing the performance of GOA-1, GOA-23 and MSIGOA. We can learn that GOA-1 is superior to GOA. This indicates that IBUF improves the performance of GOA. GOA-23 with IBUF removed is not better than MSIGOA, which also shows that IBUF is able to further improve the performance of GOA. In addition, we observe that IBUF is more effective on higher dimensional functions. In conclusion, IBUF has the capability to enhance the performance of GOA.

The effect of APTS on GOA can be assessed by analyzing the performance of GOA-2, GOA-13 and MSIGOA. We can learn that GOA-2 is superior to GOA. This indicates that APTS improves the performance of GOA. GOA-13 with APTS removed is not better than MSIGOA, which also shows that APTS is able to further improve the performance of GOA. In addition, we observe that IBUF's ranking fluctuates less, which suggests that the adaptive strategy is well adapted. In conclusion, APTS can improve the optimization ability of GOA.

The effect of DPRM on GOA can be assessed by analyzing the performance of GOA-3, GOA-12 and MSIGOA. We can learn that GOA-3 is superior to GOA. This indicates that DPRM improves the performance of GOA. The GOA-12 with DPRM removed shows a significant decrease compared to MSIGOA. In addition, we observe that GOA-13 and GOA-23 have the best rankings except for MSIGOA, which indicates a good compatibility between DPRM and other strategies. In conclusion, DPRM can improve the optimization of GOA.

**Table 6.** Results of the analysis using Friedman test

| Test suite | Dimension | GOA | GOA-1 | GOA-2 | GOA-3 | GOA-12 | GOA-13 | GOA-23 | MSIGOA | P-value |
|---|---|---|---|---|---|---|---|---|---|---|
| CEC-2017 | D=10 | 5.79 | 5.03 | 4.24 | 4.62 | 4.90 | 4.83 | 3.76 | 2.83 | 2.93E-04 |
| | D=30 | 7.45 | 4.97 | 4.93 | 4.52 | 5.90 | 3.34 | 3.14 | 1.76 | 9.40E-20 |
| | D=50 | 7.69 | 4.34 | 4.83 | 5.69 | 5.69 | 2.93 | 3.24 | 1.59 | 1.08E-23 |
| | D=100 | 7.62 | 3.31 | 5.00 | 6.59 | 5.34 | 2.34 | 3.86 | 1.93 | 3.64E-26 |
| CEC-2022 | D=10 | 5.67 | 4.42 | 4.33 | 4.58 | 4.17 | 5.00 | 5.00 | 2.83 | 2.13E-01 |
| | D=20 | 6.5 | 4.92 | 3.67 | 5.00 | 5.92 | 3.67 | 3.83 | 2.50 | 9.20E-04 |
| Total mean rank | | 6.79 | 4.50 | 4.50 | 5.17 | 5.32 | 3.69 | 3.81 | 2.24 | |

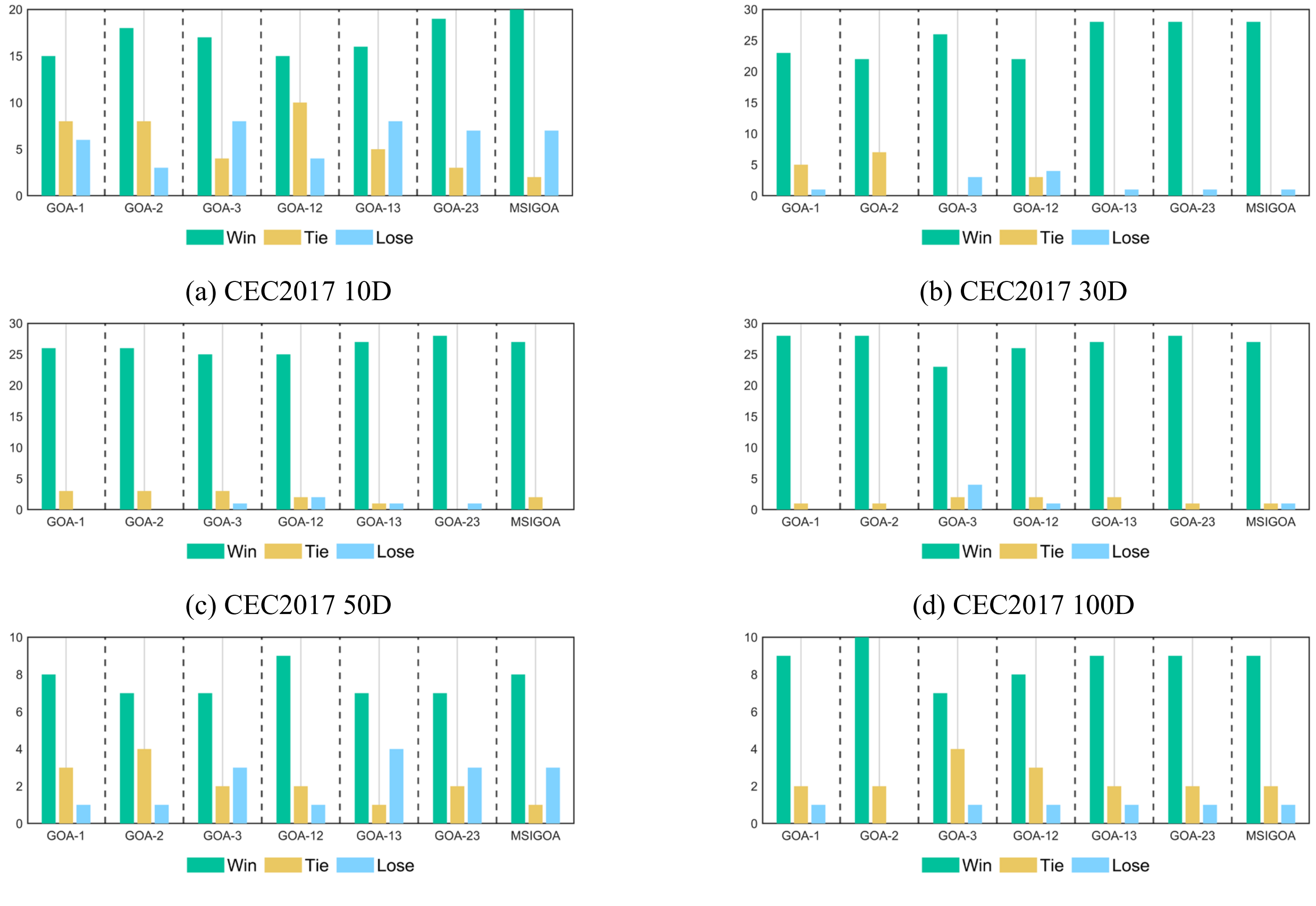

(a) CEC2017 10D (b) CEC2017 30D

(c) CEC2017 50D (d) CEC2017 100D

(e) CEC2022 10D (f) CEC2022 20D

**Figure 4.** Wilcoxon rank sum test results for the GOA variants. (a) CEC2017 10D, (b) CEC2017 30D, (c) CEC2017 50D, (d) CEC2017 100D, (e) CEC2022 10D, (f) CEC2022 20D

*4.4. Experimental results analysis on benchmark functions*

In this section, to ensure the diversity and comprehensiveness of the experiments, we selected the 50D/100D functions of CEC2017 and the 10D/20D functions of CEC2022 to measure the performance of MSIGOA and competitors. Although CEC2017 can realize the testing of low-dimensional functions, in this paper, we chose the CEC2022 function as the low-dimensional performance test object in order to evaluate the performance of MSIGOA more comprehensively.

4.4.1. Lower dimensional functions from CEC 2022

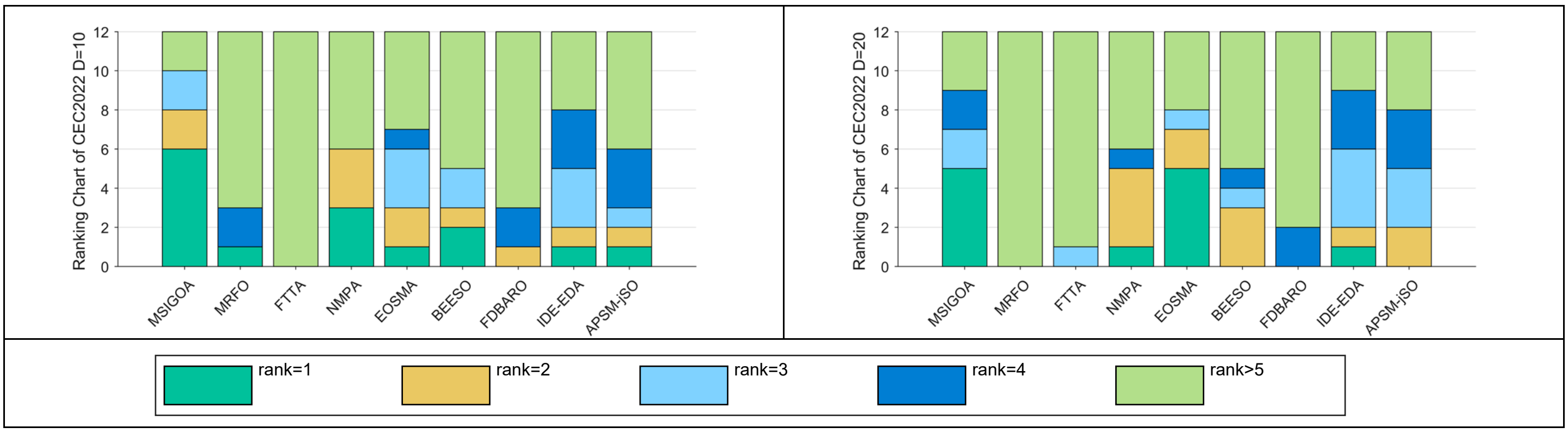

**Figure 5.** Comparison of CEC 2022 test functions rankings of MSIGOA and competitors

Table A1 and Table A2 in Appendix A recorded the average (Mean), optimal (Best) and standard deviation (Std) values obtained by MSIGOA and the competitors on the 12 benchmark functions of CEC2022 10D/20D. In general, lower mean values indicate better convergence of the algorithms, while lower standard deviation reflects more stable optimization performance, and smaller optimal values indicate better search capability. The results highlight the excellent optimization performance of MSIGOA within the CEC2020 framework, demonstrating its potential to solve real-world problems. Figure 5 shows the ranking of the different algorithms in both dimensions. To improve clarity, we categorize the rankings into five groups: Rank 1 (best), Rank 2,

Rank 3, Rank 4 and Other Rank (greater than or equal to 5). It is clear from the figure that the MSIGOA algorithm consistently leads in the average rankings for 10 and 20 dimensions. In the case of the 10-dimensional test, MSIGOA achieves the best average ranking for 6 functions, the second-best average ranking for 2 functions, and the third best average ranking for 2 functions. MSIGOA continues to be well optimized as the dimensionality increases. In the 20-dimensional test, MSIGOA still achieved optimal average rankings for 5 functions and second-best average rankings for 2 functions.

Figure 6 plots the convergence curves of parts of each algorithm while solving the CEC2022 test set to show the convergence performance of MSIGOA and the competitors. According to Figure 6, MSIGOA exhibits faster convergence and less fluctuation on most of the test functions. The comprehensive analysis shows that for functions F1 and F8, when all the compared algorithms approach the global optimal solution gradually with the increase of iterations, MSIGOA always converges at a leading speed and demonstrates the optimal performance. For F2, F3, F4, F6 and F7, MSIGOA is not the fastest convergence speed in the early stage, but due to its superior exploratory and local search ability, it gradually outperforms the other algorithms in terms of convergence speed, and ultimately reaches the optimum at the fastest time, indicating that MSIGOA has a stronger potential for global search and the ability to overcome local minima. For F5, F9, F11, F12, MSIGOA and the other algorithms involved in the comparison are similar in convergence performance. MSIGOA's good convergence is mainly due to the dominant population - based restart mechanism. Guided by the optimal individual and dominant population, it achieves fast convergence and high - precision. The excellent convergence of MSIGOA on hybrid and composite functions reflects its balance between exploitation and exploration. This is attributed to the Iteration - based updating framework (IBUF) and Adaptive parameter tuning strategy (APTS). IBUF divides the search process into stages, enabling MSIGOA to choose appropriate search methods at different stages. APTS enables smooth changes in the search range through adaptive parameter. In conclusion, Figure 6 provides comparative information on the solution efficiency of different algorithms for various types of optimization problems, demonstrating the strong convergence performance of MSIGOA on the CEC2022 test functions.

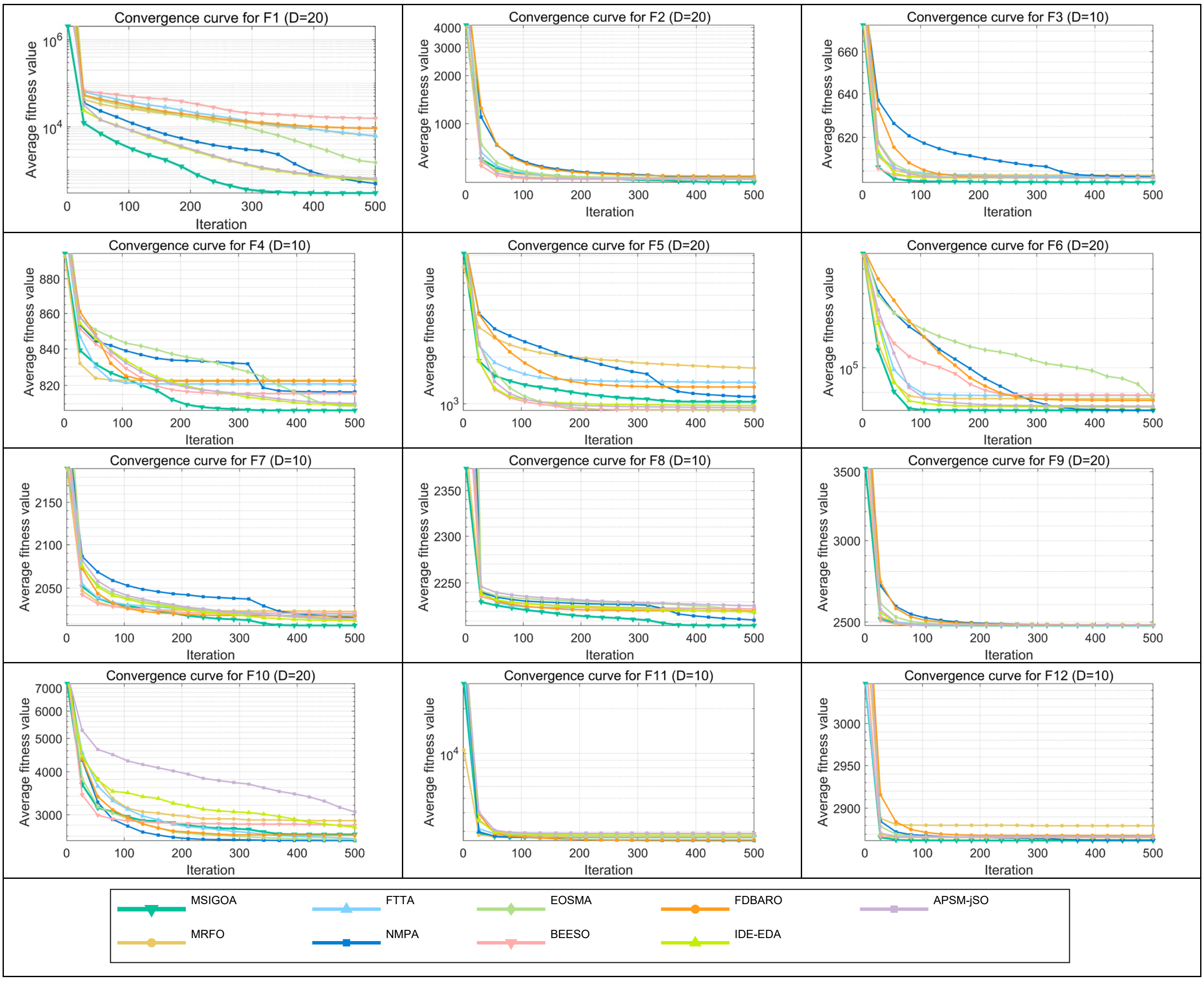

**Figure 6.** Convergence curve of MSIGOA and competitors in CEC 2022

### 4.4.2. Higher dimensional functions from CEC 2017

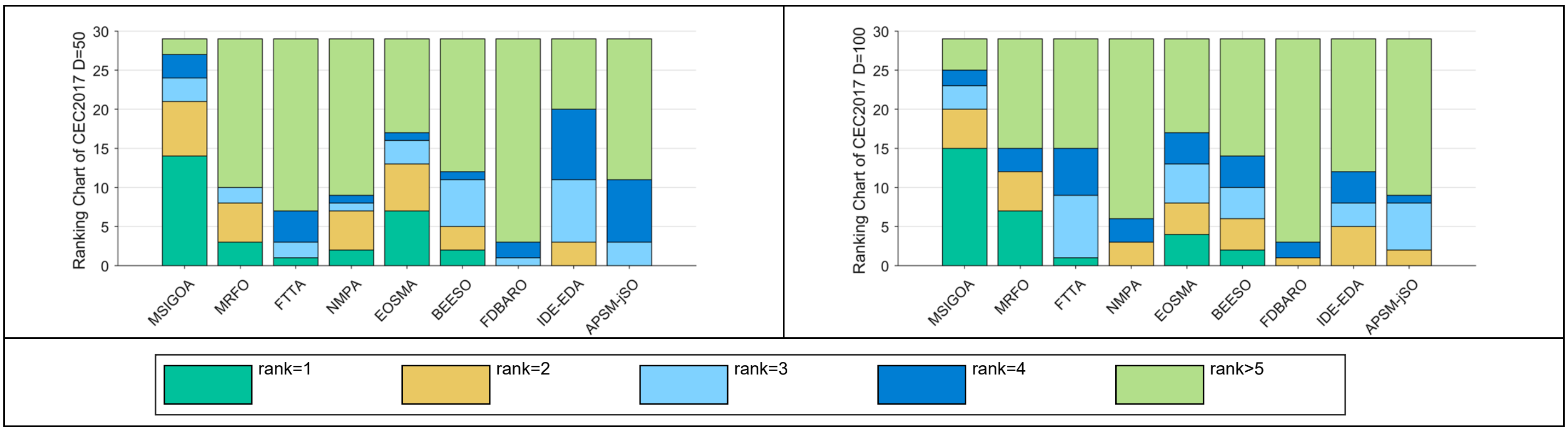

**Figure 7.** Comparison of CEC 2017 test functions rankings of MSIGOA and competitors

The test results of MSIGOA and competitors on 50- and 100-dimensional CEC2017 functions are shown in Table A3 and Table A4. To better illustrate the comparative ranking of MSIGOA with other algorithms, we used a stacked ranking graph as shown in Figure 7. As can be seen in Figure 7, MSIGOA outperforms the other methods in terms of average rankings for the 50-dimensional tests, securing the first position for 14 functions, the second position for 7 functions, and only one of the lowest rankings. In the context of the 100 dimensions, although MSIGOA's performance dropped slightly, it still managed to secure the first position for 15 functions, the second position for 5 functions, and the third position for 3 functions. MSIGOA's overall performance proves its strong scalability and effectiveness. In terms of overall ranking, MSIGOA ranks the highest among the 10 algorithms, clearly outperforming the other algorithms. Specifically, in the unimodal functions F1 and F2, MSIGOA shows excellent exploitation, ranking second and third in F1 on 50D and 100D, respectively. It ranks first under both dimensions of F2. The results show that MSIGOA has the ability to quickly solve to a more optimal solution. For the multi-peak functions F3-F9, MSIGOA is not the best performer on 50D, but shows better stability. The best results for four of these functions were achieved on 100D. These results show that MSIGOA's unique diversified search method effectively prevents it from falling into local optimality and provides robust overall performance when dealing with complex multi-peak problems. For hybrid and combinatorial functions, MSIGOA ranks in the top three on 18 functions on 50D, with only two functions ranking fifth. MSIGOA ranks in the top three on 17 functions on 100D, with only three functions ranking sixth. MSIGOA ranks in the top three on 17 functions on 100D, with only three functions ranking sixth. These results further emphasize MSIGOA's ability to provide efficient solutions to difficult problems. In conclusion, MSIGOA, with its excellent performance, provides satisfactory overall results in solving these complex problems.

The stability of the algorithm is also important for high-dimensional complex problems. To further validate the stability of MSIGOA when dealing with high-dimensional problems, the comparison results are analyzed in a box plot, as shown in Figure 8. A box plot is a statistical tool that utilizes five key statistics to illustrate the distribution of data: minimum, upper quartile, median, lower quartile, and maximum. The upper and lower boundaries of the box correspond to the third and first quartiles, respectively, while the interior contains the middle 50% of the data. The box plots help to visualize and analyze the data distribution, outliers and symmetry. Based on the results shown in Figure 8, in most cases, the box plots of MSIGOA look narrow and stay at the lowest position, which indicates superior optimization performance and stability in dealing with a variety of complex problems compared to other competitors. Since it is easy to find similar good values in the function, some box plots may show small differences, leading to smaller differences. Overall, the MSIGOA algorithm shows better stability in box plots of CEC2017 functions.

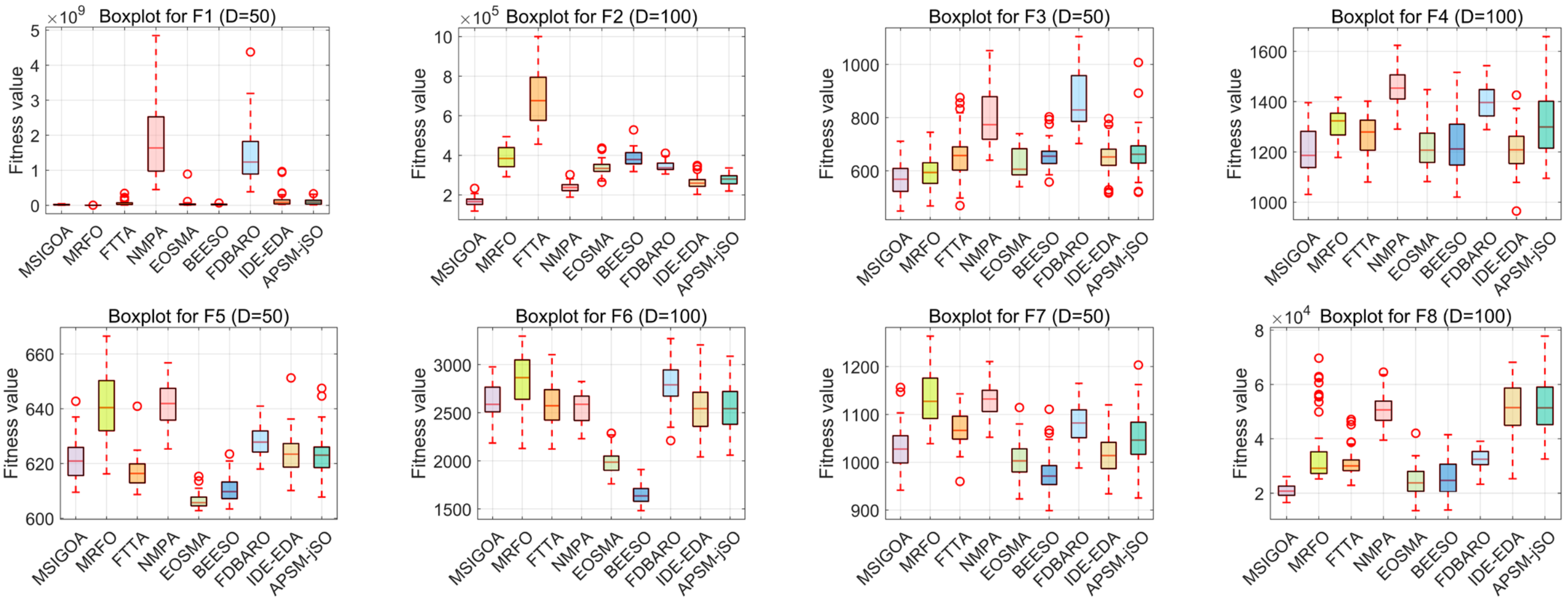

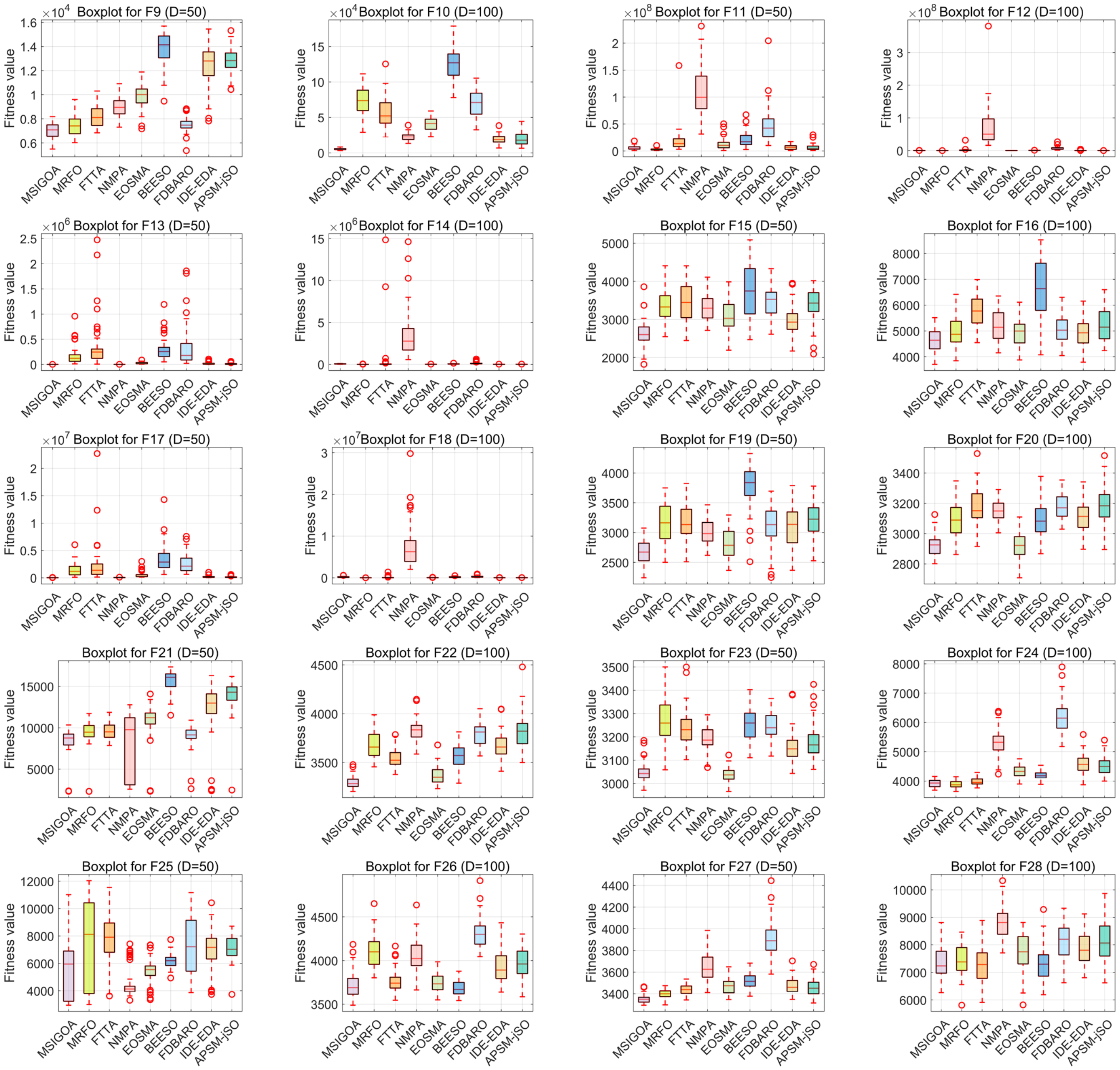

**Figure 8.** Boxplots of MSIGOA and competitors in CEC 2017

### 4.4.3. Statistical Tests

**Table 7.** Wilcoxon rank sum test statistical results of MSIGOA and competitors

| Test suite | MSIGOA v.s. +/=/- | MRFO | FTTA | NMPA | EOSMA | BEESO | FDBARO | IDE-EDA | APSM-JSO |
|---|---|---|---|---|---|---|---|---|---|
| CEC-22 | D=10 | 11/0/1 | 10/2/0 | 8/3/1 | 8/2/2 | 8/2/2 | 10/1/1 | 8/3/1 | 9/2/1 |
| | D=20 | 12/0/0 | 10/1/1 | 10/2/0 | 6/0/6 | 9/1/2 | 11/0/1 | 7/3/2 | 8/1/3 |
| CEC-17 | D=50 | 24/1/4 | 25/2/2 | 26/3/0 | 17/5/7 | 21/3/5 | 29/0/0 | 22/4/3 | 24/3/2 |
| | D=100 | 19/4/6 | 23/3/3 | 27/2/0 | 17/3/9 | 18/6/5 | 29/0/0 | 23/3/3 | 24/1/4 |

In this section, we apply the non-parametric Wilcoxon rank-sum test to compare the performance differences between MSIGOA and competitors, with relevant results displayed in Table 7. When the p-value is less than 0.05, it indicates a significant difference between MSIGOA and competing algorithms; otherwise, there is no significant difference. The symbol "+/=/-" is uti-lized to denote that MSIGOA algorithm wins, ties and loses than other competitive algorithms, respectively. The wins, ties, losses between MSIGOA and the comparison algorithms in the three test sets are visually depicted in Figure 9. The details are presented below.

For CEC 2017 50D, MSIGOA wins (loses) MRFO, FTTA, NMPA, EOSMA, BEESO, FDBARO, IDE-EDA, APSM-jSO on 24(4), 25(2), 26(0), 17(7), 21(5), 29(0), 22(3) and 24(2) test functions. That is, AMRIME beats MRFO, FTTA, NMPA, EOSMA, BEESO, FDBARO, IDE-EDA and APSM-jSO.

For CEC 2017 100D, MSIGOA wins (loses) MRFO, FTTA, NMPA, EOSMA, BEESO, FDBARO, IDE-EDA, APSM-jSO on 19(6), 23(3), 27(0), 17(9), 18(5), 29(0), 23(3) and 24(4) test functions. That is, AMRIME beats MRFO, FTTA, NMPA, EOSMA, BEESO, FDBARO, IDE-EDA and APSM-jSO.

For CEC 2022 10D, MSIGOA wins (loses) MRFO, FTTA, NMPA, EOSMA, BEESO, FDBARO, IDE-EDA, APSM-jSO on 11(1), 10(0), 8(1), 8(2), 8(2), 10(1), 8(1) and 9(1) test functions. That is, AMRIME beats MRFO, FTTA, NMPA, EOSMA, BEESO, FDBARO, IDE-EDA and APSM-jSO.

For CEC 2022 20D, MSIGOA wins (loses) MRFO, FTTA, NMPA, EOSMA, BEESO, FDBARO, IDE-EDA, APSM-jSO on 12(0), 10(1), 10(0), 6(6), 9(2), 11(1), 7(2) and 8(3) test functions. That is, AMRIME beats MRFO, FTTA, NMPA, EOSMA, BEESO, FDBARO, IDE-EDA and APSM-jSO.

The experimental data clearly show that MSIGOA achieves more “+” signs than the total number of “-” signs in all three test suites, indicating significant performance differences between MSIGOA and other algorithms. Based on the above analysis, the MSIGOA algorithm exhibits unique advantages compared to other competitive algorithms and excels in terms of overall performance.

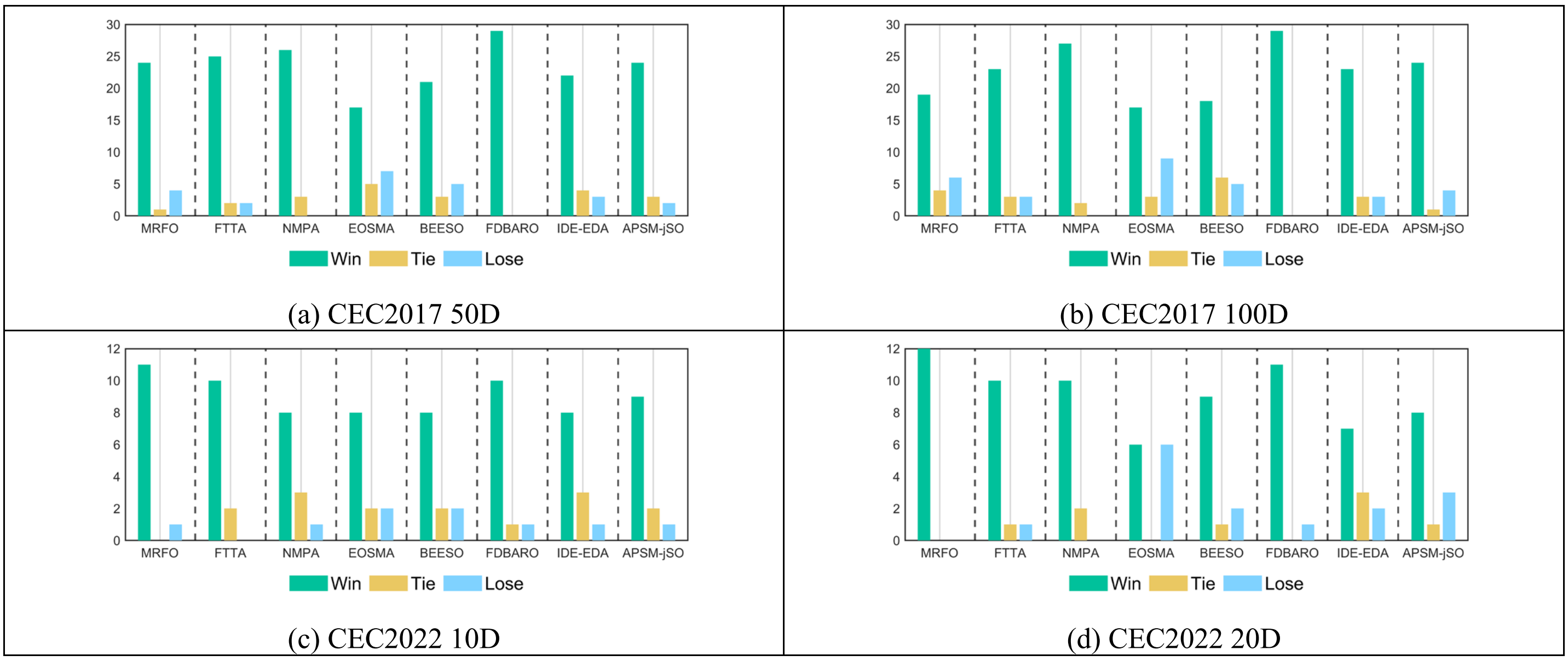

**Figure 9.** Wilcoxon rank sum test results for MSIGOA and competitors. (a)CEC2017 50D, (b)CEC2017 100D, (c)CEC2022 10D, (d)CEC2022 20D

**Table 8.** Friedman test statistical results of MSIGOA and competitors

| Test suite | Dimension | MSIGOA | MRFO | FTTA | NMPA | EOSMA | BEESO | FDBARO | IDE-EDA | APSM-JSO | P-value |
|---|---|---|---|---|---|---|---|---|---|---|---|
| CEC-17 | D=50 | 2.03 | 5.59 | 6.24 | 5.76 | 3.52 | 5.59 | 6.86 | 4.17 | 5.24 | 5.83E-12 |
| | D=100 | 2.31 | 4.38 | 4.86 | 6.72 | 3.69 | 5.24 | 7.21 | 4.90 | 5.69 | 9.30E-12 |
| CEC-22 | D=10 | 2.29 | 7.08 | 7.42 | 3.75 | 3.83 | 5.13 | 5.83 | 4.21 | 5.46 | 2.24E-05 |
| | D=20 | 2.83 | 7.50 | 6.58 | 4.83 | 3.00 | 5.33 | 6.83 | 3.58 | 4.50 | 1.09E-05 |
| Total mean rank | | 2.37 | 6.14 | 6.28 | 5.27 | 3.51 | 5.32 | 6.68 | 4.22 | 5.22 | |

In this section, we apply the nonparametric Friedman test to rank the performance of the MSIGOA algorithm and other competitive algorithms on the three test sets, with specific results presented in Table 8. Across different cases, the average rankings of the MSIGOA are 2.03, 2.31, 2.29 and 2.83, respectively, all ranking first. Its overall average ranking is 2.37, which also secures the top spot. Furthermore, Figure 10 exhibits the Friedman rankings of MSIGOA and its comparison algorithms for different test sets. These statistical results fully demonstrate the superiority of the MSIGOA in overall performance and highlight its exceptional capabilities. The detailed analysis is as follows.

For CEC 2017 50D, MSIGOA is ranked first followed by EOSMA, IDE-EDA, APSM-jSO, BEESO/MRFO, NMPA, FTTA and FDBARO. That is, MSIGOA outperforms the 8 comparison algorithms on CEC 2017 50D.

For CEC 2017 100D, MSIGOA is ranked first followed by EOSMA, MRFO, FTTA, IDE-EDA, BEESO, APSM-JSO, NMPA and FDBARO. That is, MSIGOA outperforms the 8 comparison algorithms on CEC 2017 100D.

For CEC 2022 10D, MSIGOA is ranked first followed by NMPA, EOSMA, IDE-EDA, BEESO, APSM-JSO, FDBARO, MRFO and FTTA. That is, MSIGOA outperforms the 8 comparison algorithms on CEC 2022 10D.

For CEC 2022 20D, MSIGOA is ranked first followed by EOSMA, IDE-EDA, APSM-jSO, NMPA, BEESO, FTTA, FDBARO and MRFO. That is, MSIGOA outperforms the 8 comparison algorithms on CEC 2022 20D.

In conclusion, the performance of our proposed MSIGOA is summarized as the best under both statistical test methods.

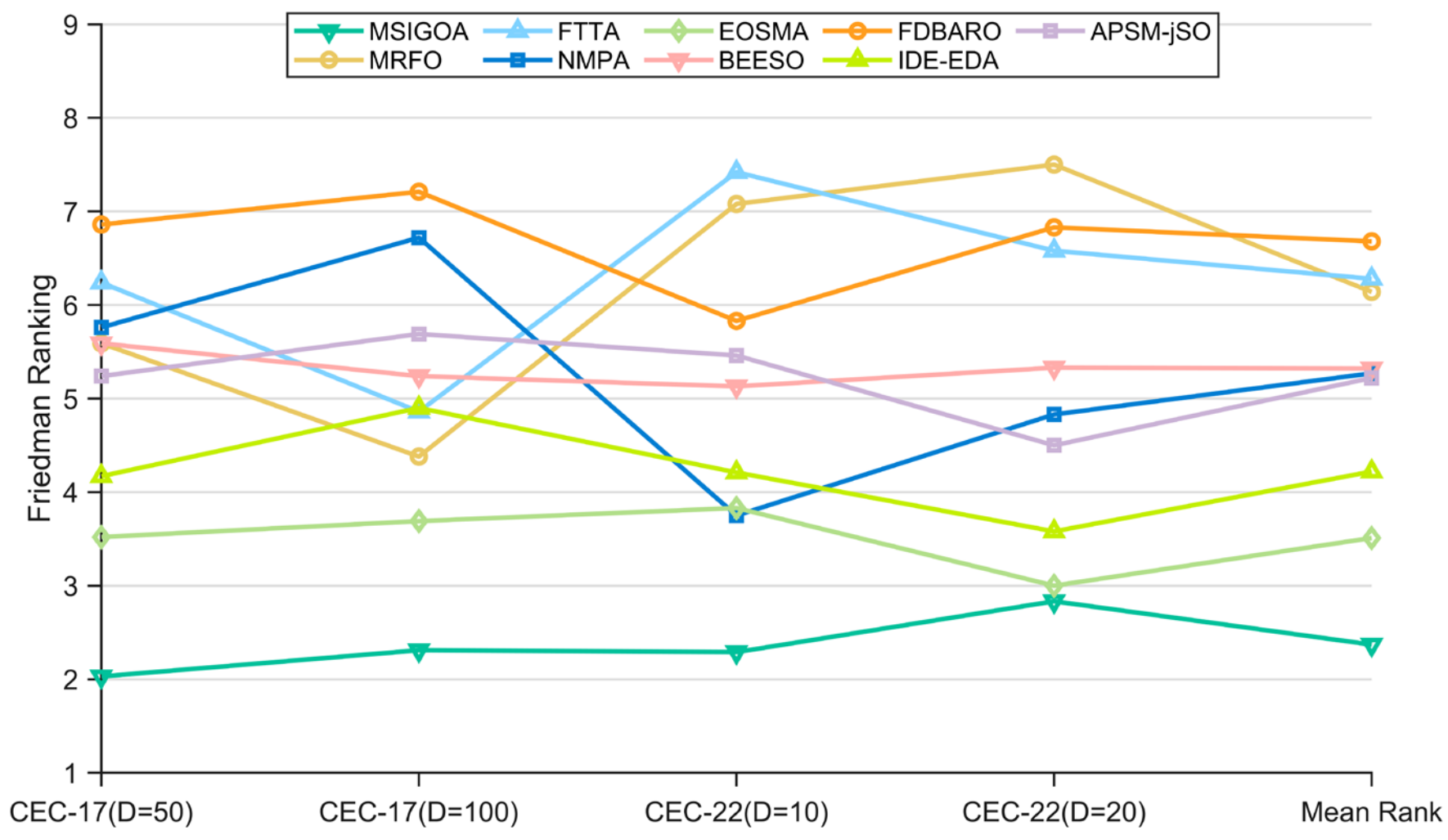

**Figure 10.** Friedman ranking of MSIGOA and competitors

*4.5. Experimental results analysis on engineering design optimization problems*

This subsection evaluates the effectiveness of the MSIGOA algorithm in addressing engineering design problems, specifically constrained optimization challenges. A total of six engineering problems have been used for this assessment: the welded beam design problem, the pressure vessel design problem and the tension/compression spring design problem. Due to the numerous inequality constraints in these problems, if any of these constraints are violated, algorithms typically employ a penalty function to achieve a feasible solution. The results obtained by MSIGOA are compared with those produced by other algorithms to demonstrate its efficacy.

4.5.1. Tension/compression spring design problem

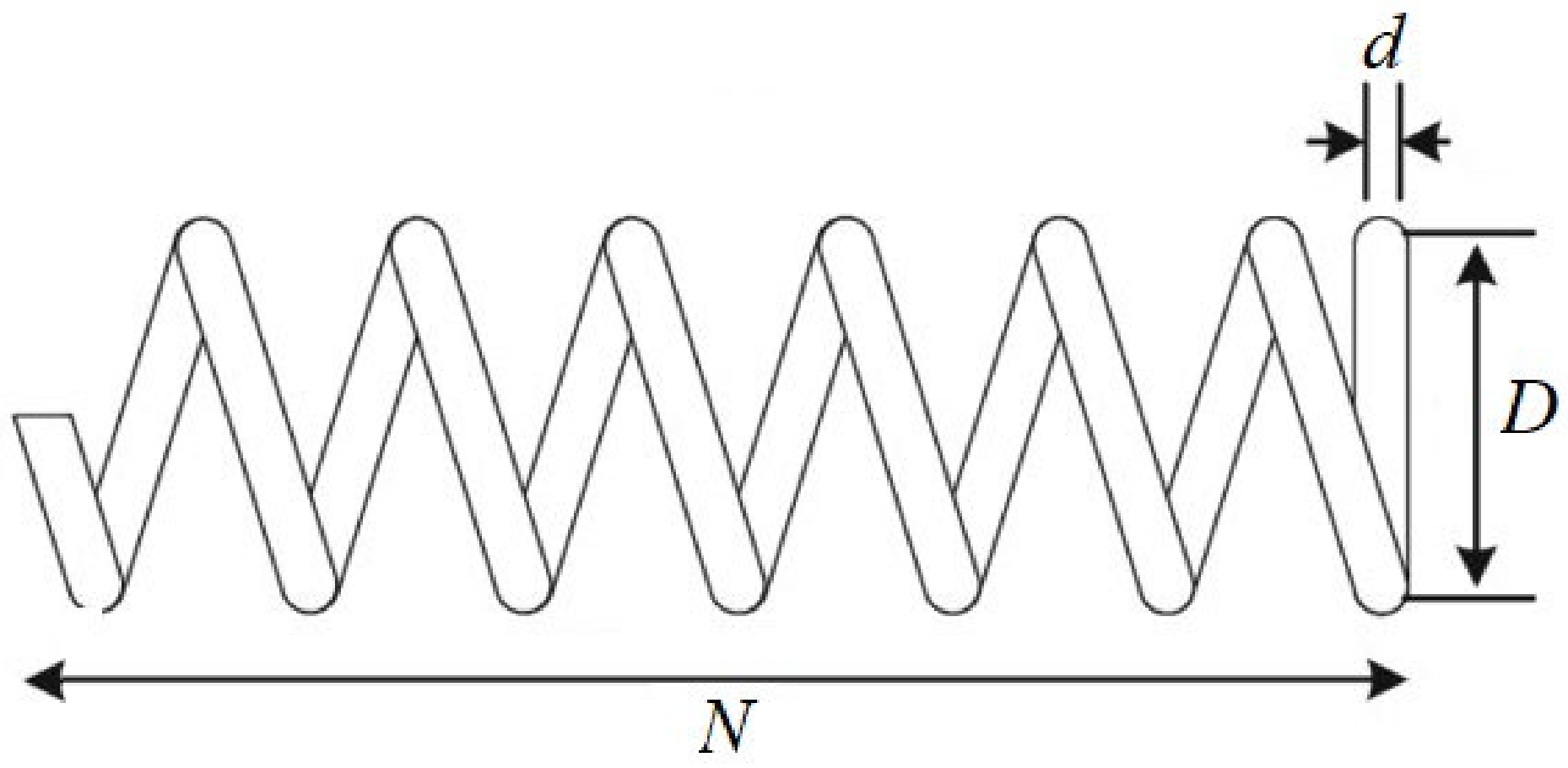

**Figure 11.** Structure of tension/compression spring

Springs are elastic objects, often made of metal wire or other elastic materials, and they have extensive importance and applications in both industrial production and daily life. The spring design problem for tension/compression springs aims to minimize material consumption and weight while meeting design requirements such as minimal deflection, shear stress, resonance frequency, and outside diameter limit. A spring is represented as shown in Figure 11, where *d* represents wire diameter, *D* represents mean coil diameter, and *N* represents the number of effective coils. The mathematical model for the spring design optimization problem is as follows:
Variables:

$$x = [x_1, x_2, x_3] = [d, D, N] \tag{15}$$

Minimize:

$$f(x) = (x_3 + 2)x_2 x_1^2 \tag{16}$$

Subject to:

$$\begin{aligned} g_1(x) &= 1 - \frac{x_2^3 x_3}{71785 x_1^4} \le 0, \\ g_2(x) &= \frac{4x_2^2 - x_1 x_2}{12566(x_2 x_1^3 - x_1^4)} + \frac{1}{5108 x_1^2} \le 0, \\ g_3(x) &= 1 - \frac{140.45 x_1}{x_2^2 x_3} \le 0, \\ g_4(x) &= \frac{x_1 + x_2}{1.5} - 1 \le 0 \end{aligned} \tag{17}$$

With bounds:

$$0.05 \le x_1 \le 2.00,\ 0.25 \le x_2 \le 1.30,\ 2.00 \le x_3 \le 15.0 \tag{18}$$

Table 9 presents the optimization results for the tension/compression spring design variables obtained using various optimization algorithms. The table displays the optimal values for the design variables (*d*, *D*, and *N*) alongside the corresponding optimum cost achieved by each algorithm. Notably, the MSIGOA algorithm demonstrates superior performance, yielding optimal design variable values of *d* = 0.05168906, *D* = 0.35671774, and *N* = 11.288965, with an associated optimum cost of 0.01266523. This outcome MSIGOA as the most effective algorithm among the tested methods for optimizing the tension/compression spring design problems.

**Table 9.** Experimental Results of MSIGOA in design of tension/compressive springs

| Algorithm | Optimal values for Variables | | | Optimal value |
|---|---|---|---|---|
| | *d/m* | *D/m* | *N/m* | |
| MSIGOA | 0.05168906 | 0.35671774 | 11.288965 | 0.01266523 |
| GOA | 0.05460254 | 0.43026821 | 8.0148045 | 0.01284717 |
| MRFO | 0.05272665 | 0.38191964 | 9.9862305 | 0.01272667 |
| FTTA | 0.05244573 | 0.37475123 | 10.329422 | 0.01270884 |
| NMPA | 0.05332871 | 0.39739971 | 9.2642454 | 0.01273068 |
| EOSMA | 0.05297470 | 0.38832553 | 9.6552854 | 0.01270152 |
| BEESO | 0.05188524 | 0.36144834 | 11.017686 | 0.01266683 |
| FDBARO | 0.05160989 | 0.35481601 | 11.401335 | 0.01266535 |
| IDE-EDA | 0.05185005 | 0.36060303 | 11.064787 | 0.01266572 |
| APSM-jSO | 0.05144754 | 0.35093265 | 11.636509 | 0.01266649 |

4.5.2. Pressure vessel design problem

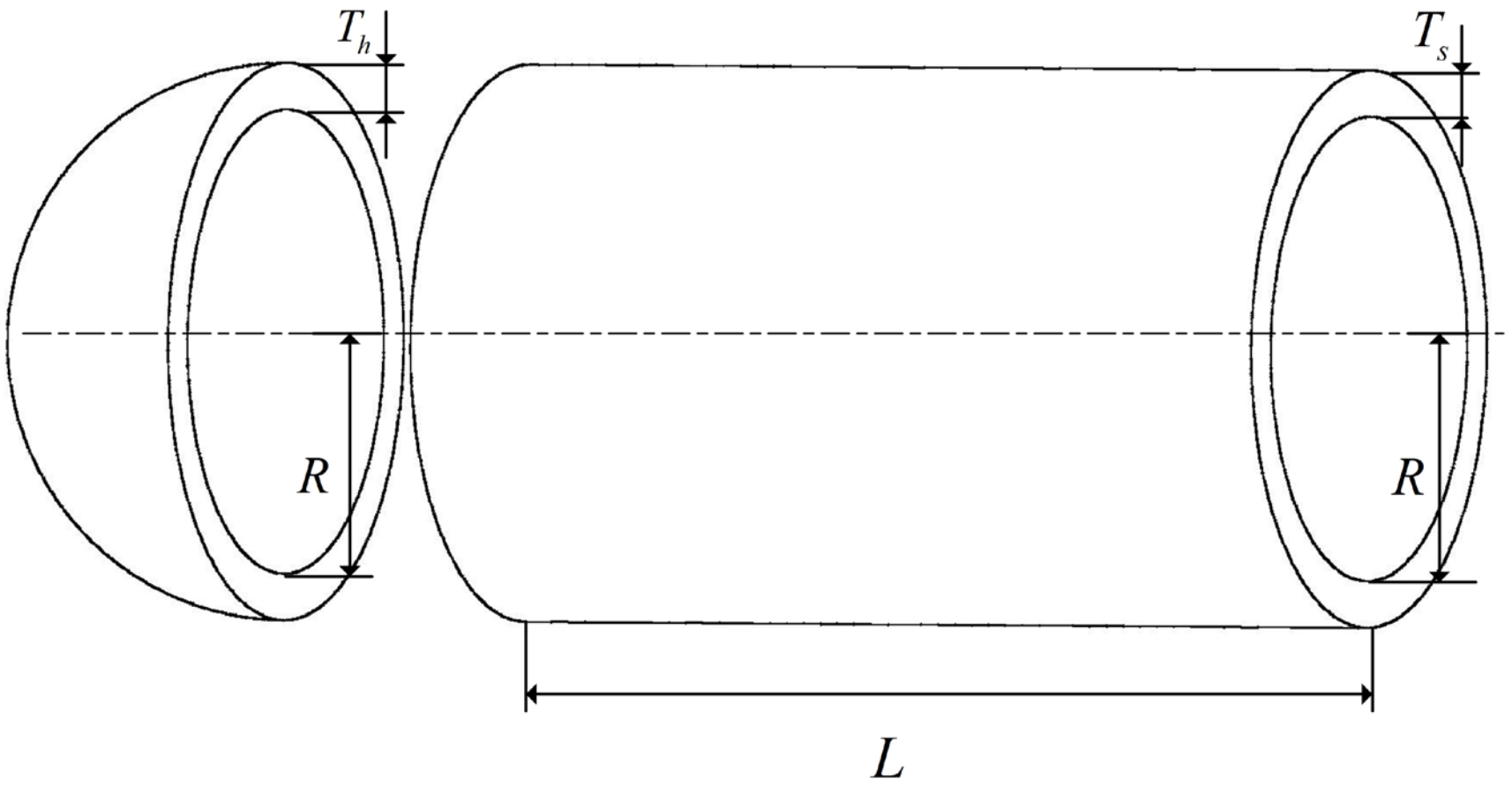

**Figure 12.** Structure of pressure vessel

Figure 12 illustrates a general scheme of the pressure vessel design problem. This problem aims to minimize the cost of used materials, welding, and forming by finding the optimum value for the four variables: (Ts) Thickness of the Shell, (Th) Thickness of the head, (R) inner radius, and (L) The length of the cylindrical section without considering the head. Equation (19-22) shows the formal mathematical description of the problem.
Variables:

$$x = \left[x_1,\ x_2,\ x_3, x_4\right] = \left[T_s,\ T_h,\ R, L\right] \tag{19}$$

Minimize:

$$f(x) = 0.6224 x_1 x_3 x_4 + 1.7781 x_2 x_3^2 + 3.1661 x_1^2 x_4 + 19.84 x_1^2 x_3 \tag{20}$$

Subject to:

$$\begin{aligned} & g_1(x) = x_1 + 0.0193 x_3 \le 0 \\ & g_2(x) = x_2 + 0.00954 x_3 \le 0 \\ & g_3(x) = -\pi x_3^2 x_4 + \frac{4}{3}\pi x_3^3 + 1296000 \le 0 \\ & g_4(x) = x_4 - 240 \le 0 \end{aligned} \tag{21}$$

With bounds:

$$0.1 \le x_1, x_4 \le 2.0, 0.1 \le x_2, x_3 \le 10 \tag{22}$$

Table 10 presents the optimization results for the pressure vessel design problem, encompassing four key design variables: Ts, Th, R, and L. Each algorithm's performance is evaluated in terms of the optimal values achieved for these variables and the corresponding optimum cost function values. Among the algorithms examined, the MSIGOA algorithm demonstrates notable effectiveness by yielding optimal design variable values of Ts = 0.77816864, Th = 0.38464916, R = 40.3196187, and L = 200, with a resulting optimum cost of 5885.33277. This outcome underscores MSIGOA's capability to efficiently find solutions that minimize the cost function for the pressure vessel design.

**Table 10.** Experimental Results of MSIGOA in design of pressure vessel design

| Algorithm | Optimal values for Variables | | | | Optimal value |
|---|---|---|---|---|---|
| | $T_s/m$ | $T_h/m$ | $R/m$ | $L/m$ | |
| MSIGOA | 0.77816864 | 0.38464916 | 40.3196187 | 200.0000000 | 5885.33277 |
| GOA | 0.77896429 | 0.39045184 | 40.3348099 | 199.7968246 | 5906.01803 |
| MRFO | 0.77927028 | 0.38519906 | 40.3350275 | 199.8876156 | 5895.02730 |
| FTTA | 0.81758383 | 0.4072927 | 42.0917457 | 176.7196352 | 6000.46259 |
| NMPA | 0.7785454 | 0.3875910 | 40.3324246 | 199.9429427 | 5896.42811 |
| EOSMA | 0.7931381 | 0.3932639 | 40.6711482 | 196.2178371 | 5994.18572 |
| BEESO | 0.78825247 | 0.38979723 | 40.8353571 | 193.0706748 | 5906.97945 |
| FDBARO | 0.77953534 | 0.38594539 | 40.3335133 | 200.0000000 | 5901.27507 |
| IDE-EDA | 0.77823263 | 0.38495491 | 40.3222291 | 199.9905697 | 5886.90655 |
| APSM-jSO | 0.77838934 | 0.38485754 | 40.3304808 | 199.8523002 | 5886.14837 |

4.5.3. Welded beam design problem

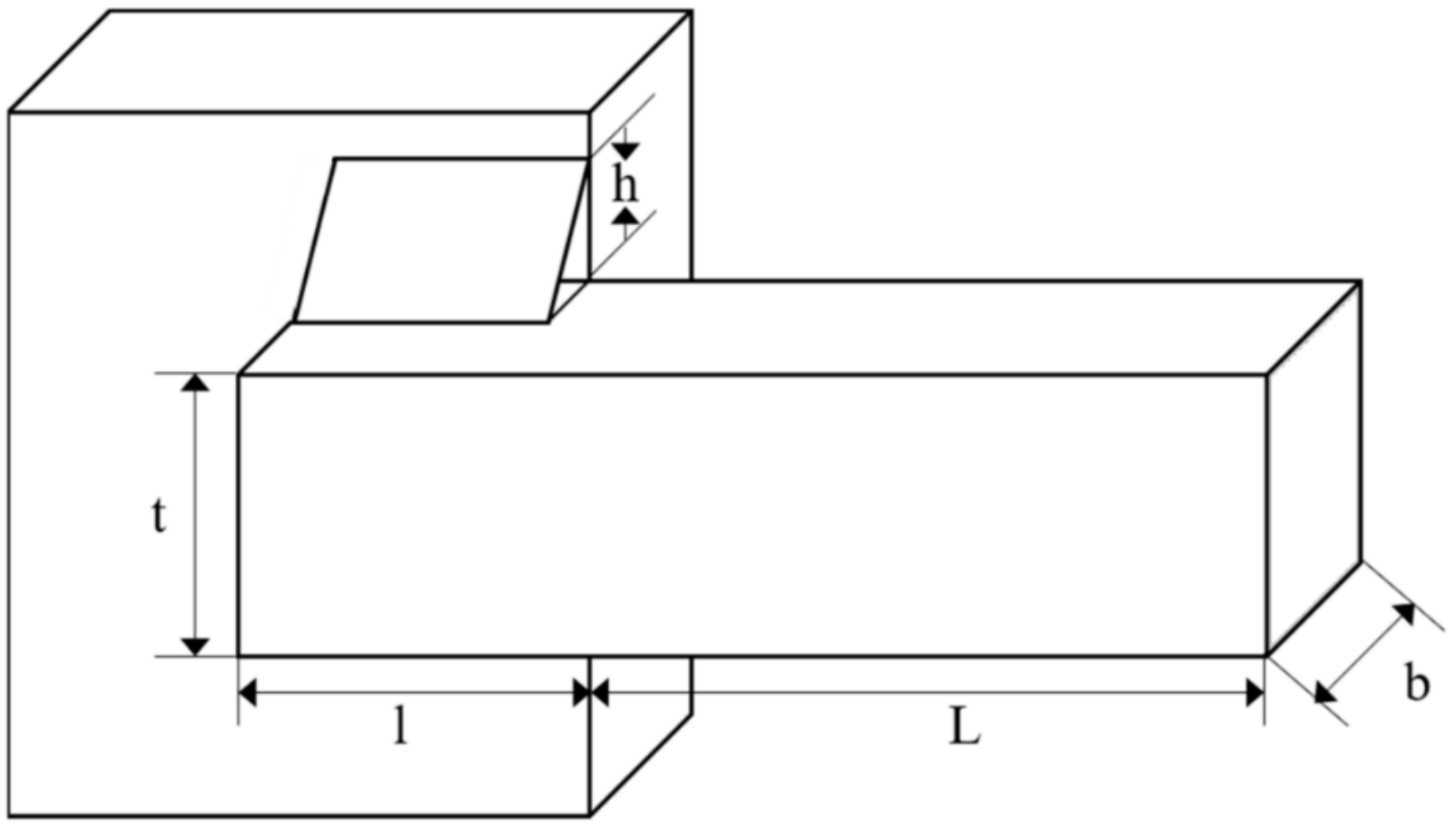

**Figure 13.** Structure of welded beam

The main objective of the welded beam design problem is to create a beam at the lowest possible cost while complying with specific constraints. Figure 13 illustrates a welded beam design where beam A is connected to element B by a welding process. This design problem comprises five non-linear inequality constraints and four decision variables. The design parameters are weld thickness h (x1), weld joint length l (x2), element width t (x3) and element thickness b (x4).
Variables:

$$x = [x_1, x_2, x_3, x_4] = [h, l, t, b] \tag{23}$$

Minimize:

$$f(x) = 0.04811 x_3 x_4 (x_2 + 14) + 1.10471 x_1^2 x_2 \tag{24}$$

Subject to:

$$\begin{aligned} g_1(x) &= x_1 - x_4 \le 0 \\ g_2(x) &= \delta(x) - \delta_{max} \le 0 \\ g_3(x) &= P - P_c(x) \le 0 \\ g_4(x) &= \tau(x) - \tau_{max} \le 0 \end{aligned} \tag{25}$$

With bounds:

$$1 \le x_1, x_2 \le 99, 10 \le x_3, x_4 \le 200 \tag{26}$$

Table 11 summarizes the performance of various algorithms when applied to the welded beam design problem. The table provides the optimum values for the design variables (h, l, t, and b) as well as the optimum value achieved by each algorithm. These results highlight that the MSIGOA algorithm achieved the optimal solution with the lowest optimum value of 1.692768264 among all the algorithms tested.

**Table 11.** Experimental Results of MSIGOA in design of welded beam design problem

| Algorithm | Optimal values for Variables | | | | Optimal value |
|---|---|---|---|---|---|
| | *h/m* | *l/m* | *t/m* | *b/m* | |
| MSIGOA | 0.2057296 | 3.2349193 | 9.0366239 | 0.2057296 | 1.6927682 |
| GOA | 0.2036878 | 3.5284673 | 9.0042333 | 0.2072416 | 1.7353445 |
| MRFO | 0.2057296 | 3.4704899 | 9.03662398 | 0.2057296 | 1.7248525 |
| FTTA | 0.205732 | 3.470471 | 9.036572 | 0.2057328 | 1.7248626 |
| NMPA | 0.1937412 | 3.5110487 | 9.0456056 | 0.2058641 | 1.7143828 |
| EOSMA | 0.1953904 | 3.4510984 | 9.0334213 | 0.2060627 | 1.7083729 |
| BEESO | 0.2054496 | 3.2594453 | 9.0413576 | 0.2057420 | 1.6965958 |
| FDBARO | 0.2056792 | 3.25655193 | 9.034199 | 0.2058606 | 1.6962085 |
| IDE-EDA | 0.2025595 | 3.3168653 | 9.0327971 | 0.2059067 | 1.6998632 |
| APSM-jSO | 0.2054688 | 3.2589378 | 9.0384377 | 0.2058823 | 1.6960441 |

## 5. Conclusions

To address the problems of unbalanced exploration and exploitation, slow convergence, easy to fall into local optimums, and low convergence accuracy of the gazelle optimization algorithm, we propose an enhanced variant of GOA called MSIGOA. We propose a new updating framework, IBUF, to ameliorate the imbalance of GOA. IBUF emphasizes the proper searching at different stages and ensures balanced transition from exploration to exploitation. Two adaptive parameter control strategies further enrich the adaptive capability of MSIGOA, enabling it to avoid premature convergence and effectively explore different regions of the solution space. The dominant population-based restart mechanism enhances the population diversity and effectively improves the quality of candidate solutions. The performance of MSIGOA is well validated against other basic algorithms, excellent algorithmic variants, and classical GOA in CEC2017, CEC2022, and engineering design optimization problems. Numerous test results prove that MSIGOA outperforms basic GOA and other competitors. It achieves faster convergence, higher solution quality, and stronger robustness in different optimization environments, making it particularly suitable for real-world applications that require a balance between global exploration and local exploitation. The MSIGOA algorithm generally performs well but fails to provide the best and most stable solutions for some functions in the CEC2017 and CEC2022 test sets. Additionally, it consumes more time solving complex optimization problems. This indicates there is still room for improvement in the algorithm's structure and search mechanism. Furthermore, the performance evaluation of MSIGOA can be analyzed by more comprehensive experiments, such as using the GKLS function generator.

In the future, MSIGOA will show promising potential by addressing specific challenges in the fields of machine learning, medicine, finance and automation. In addition, exploring the capabilities of MSIGOA in multi-objective optimization scenarios, where conflicting objectives need to be optimized simultaneously, provides an opportunity to develop hybrid strategies that integrate multi-objective techniques. In addition, investigating hybridization with other advanced meta-heuristic algorithms and machine learning methods is expected to synergistically exploit strengths and mitigate weaknesses, thereby improving optimization performance. In conclusion, it is hoped that MSIGOA can be developed into a versatile and powerful optimization tool for real world optimization problems.

## Competing Interest

The authors declare that the authors have no competing interests as defined by Nature Research, or other interests that might be perceived to influence the results and/or discussion reported in this paper.

## Data Availability Statement

The data is provided within the manuscript.

## Conflicts of Interest

The authors declare no conflict of interest.

**Acknowledgements:** This work was supported by the Scientific Research Fund of Zhejiang Provincial Education Department (Grant No. Y202456182).

## Author contributions

**Qi Diao**: conceptualization, methodology, writing, data testing, reviewing, software, supervision, formal analysis. **Chengyue Xie:** writing, data testing, reviewing. **Yuchen Yin**: methodology, writing, reviewing. **Hoileong Lee**: reviewing, formal analysis. **Haolong Yang:** reviewing, formal analysis.

## Appendix A

**Table A1.** Experimental results of MSIGOA and competitors in CEC 2022 (D=10)

| Function | Index | MSGOA | MRFO | FTTA | NMPA | EOSMA | BEESO | FDBARO | IDE-EDA | APSM-JSO |
|---|---|---|---|---|---|---|---|---|---|---|
| | Best | 3.0000E+02 | 3.0000E+02 | 3.0000E+02 | 3.0000E+02 | 3.0000E+02 | 3.0007E+02 | 3.0005E+02 | 3.0000E+02 | 3.0000E+02 |
| F1 | Mean | 3.0000E+02 | 3.0000E+02 | 3.0417E+02 | 3.0000E+02 | 3.0002E+02 | 3.3296E+02 | 3.1934E+02 | 3.0000E+02 | 3.0000E+02 |
| | Std | 8.0389E-15 | 6.5010E-04 | 2.2155E+01 | 1.1441E-03 | 2.9271E-02 | 3.5884E+01 | 6.2458E+01 | 3.5864E-10 | 3.9518E-11 |
| | Best | 4.0000E+02 | 4.0000E+02 | 4.0001E+02 | 4.0000E+02 | 4.0001E+02 | 4.0002E+02 | 4.0000E+02 | 4.0000E+02 | 4.0000E+02 |
| F2 | Mean | 4.0255E+02 | 4.0896E+02 | 4.1553E+02 | 4.0018E+02 | 4.0322E+02 | 4.0523E+02 | 4.0378E+02 | 4.0399E+02 | 4.0557E+02 |
| | Std | 3.1443E+00 | 1.8848E+01 | 2.3366E+01 | 8.4521E-01 | 3.3927E+00 | 3.0541E+00 | 3.2729E+00 | 3.8629E+00 | 3.2141E+00 |
| | Best | 6.0000E+02 | 6.0000E+02 | 6.0000E+02 | 6.0018E+02 | 6.0000E+02 | 6.0000E+02 | 6.0000E+02 | 6.0000E+02 | 6.0000E+02 |
| F3 | Mean | 6.0001E+02 | 6.0109E+02 | 6.0046E+02 | 6.0047E+02 | 6.0000E+02 | 6.0001E+02 | 6.0003E+02 | 6.0003E+02 | 6.0007E+02 |
| | Std | 1.9480E-02 | 1.9247E+00 | 4.6863E-01 | 1.9126E-01 | 1.7174E-03 | 1.1521E-02 | 3.8791E-02 | 2.1600E-01 | 4.5062E-01 |
| | Best | 8.0099E+02 | 8.0497E+02 | 8.0696E+02 | 8.0498E+02 | 8.0199E+02 | 8.0298E+02 | 8.0597E+02 | 8.0199E+02 | 8.0226E+02 |
| F4 | Mean | 8.0650E+02 | 8.2074E+02 | 8.1876E+02 | 8.1443E+02 | 8.0690E+02 | 8.1360E+02 | 8.2029E+02 | 8.0761E+02 | 8.0819E+02 |
| | Std | 3.1423E+00 | 9.6225E+00 | 8.9994E+00 | 5.5120E+00 | 2.7712E+00 | 7.3101E+00 | 7.4165E+00 | 3.9838E+00 | 3.2436E+00 |
| | Best | 9.0000E+02 | 9.0000E+02 | 9.0054E+02 | 9.0000E+02 | 9.0000E+02 | 9.0000E+02 | 9.0000E+02 | 9.0000E+02 | 9.0000E+02 |
| F5 | Mean | 9.0026E+02 | 9.2531E+02 | 9.3169E+02 | 9.0095E+02 | 9.0008E+02 | 9.0002E+02 | 9.0321E+02 | 9.0021E+02 | 9.0010E+02 |
| | Std | 4.7813E-01 | 7.3156E+01 | 3.6357E+01 | 2.6502E+00 | 1.6359E-01 | 4.2694E-02 | 6.4241E+00 | 6.0213E-01 | 2.2026E-01 |
| | Best | 1.8000E+03 | 1.8242E+03 | 1.8538E+03 | 1.8006E+03 | 1.8211E+03 | 1.8964E+03 | 1.8101E+03 | 1.8003E+03 | 1.8011E+03 |
| F6 | Mean | 1.8006E+03 | 2.6501E+03 | 4.2866E+03 | 1.8027E+03 | 2.0228E+03 | 4.6811E+03 | 2.2260E+03 | 1.8105E+03 | 1.8130E+03 |
| | Std | 6.5443E-01 | 1.0923E+03 | 2.0575E+03 | 1.2306E+00 | 1.9822E+02 | 2.0843E+03 | 8.9121E+02 | 9.8522E+00 | 1.0896E+01 |
| | Best | 2.0000E+03 | 2.0010E+03 | 2.0002E+03 | 2.0014E+03 | 2.0000E+03 | 2.0016E+03 | 2.0000E+03 | 2.0000E+03 | 2.0000E+03 |
| F7 | Mean | 2.0079E+03 | 2.0216E+03 | 2.0187E+03 | 2.0145E+03 | 2.0117E+03 | 2.0194E+03 | 2.0165E+03 | 2.0115E+03 | 2.0137E+03 |
| | Std | 9.1095E+00 | 8.4165E+00 | 5.1377E+00 | 8.8508E+00 | 9.8253E+00 | 9.0999E+00 | 9.8918E+00 | 9.6070E+00 | 9.4736E+00 |
| | Best | 2.2000E+03 | 2.2006E+03 | 2.2041E+03 | 2.2022E+03 | 2.2019E+03 | 2.2015E+03 | 2.2001E+03 | 2.2006E+03 | 2.2006E+03 |
| F8 | Mean | 2.2053E+03 | 2.2206E+03 | 2.2206E+03 | 2.2089E+03 | 2.2161E+03 | 2.2205E+03 | 2.2185E+03 | 2.2176E+03 | 2.2238E+03 |
| | Std | 8.4664E+00 | 6.9340E+00 | 3.1925E+00 | 7.6879E+00 | 8.1365E+00 | 6.5250E+00 | 5.6333E+00 | 9.9628E+00 | 2.4300E+01 |
| | Best | 2.5293E+03 | 2.5293E+03 | 2.5293E+03 | 2.5293E+03 | 2.5293E+03 | 2.5293E+03 | 2.5293E+03 | 2.5293E+03 | 2.5293E+03 |
| F9 | Mean | 2.5293E+03 | 2.5350E+03 | 2.5293E+03 | 2.5293E+03 | 2.5293E+03 | 2.5293E+03 | 2.5293E+03 | 2.5293E+03 | 2.5293E+03 |
| | Std | 0.0000E+00 | 2.8804E+01 | 3.9352E-02 | 4.1705E-05 | 1.4115E-08 | 4.3618E-13 | 2.4302E-03 | 1.5753E-13 | 3.5807E-13 |
| | Best | 2.5001E+03 | 2.5002E+03 | 2.5003E+03 | 2.5002E+03 | 2.5002E+03 | 2.4103E+03 | 2.5003E+03 | 2.5002E+03 | 2.5002E+03 |
| F10 | Mean | 2.5258E+03 | 2.5025E+03 | 2.5551E+03 | 2.5027E+03 | 2.5282E+03 | 2.5146E+03 | 2.5168E+03 | 2.5434E+03 | 2.5456E+03 |
| | Std | 4.6492E+01 | 1.4787E+01 | 6.1047E+01 | 1.6265E+01 | 4.8196E+01 | 4.2769E+01 | 4.1281E+01 | 5.4153E+01 | 5.4702E+01 |
| | Best | 2.6000E+03 | 2.6000E+03 | 2.6000E+03 | 2.6002E+03 | 2.6000E+03 | 2.6000E+03 | 2.6000E+03 | 2.6000E+03 | 2.6000E+03 |
| F11 | Mean | 2.7412E+03 | 2.7517E+03 | 2.8027E+03 | 2.6004E+03 | 2.8030E+03 | 2.8824E+03 | 2.6236E+03 | 2.8846E+03 | 2.9053E+03 |
| | Std | 1.5123E+02 | 1.4891E+02 | 1.3169E+02 | 1.8183E-01 | 1.4014E+02 | 7.1286E+01 | 7.5745E+01 | 6.6724E+01 | 1.1231E+02 |
| | Best | 2.8586E+03 | 2.8633E+03 | 2.8602E+03 | 2.8586E+03 | 2.8594E+03 | 2.8614E+03 | 2.8631E+03 | 2.8614E+03 | 2.8614E+03 |
| F12 | Mean | 2.8627E+03 | 2.8778E+03 | 2.8646E+03 | 2.8613E+03 | 2.8642E+03 | 2.8641E+03 | 2.8664E+03 | 2.8646E+03 | 2.8649E+03 |
| | Std | 1.7516E+00 | 1.6502E+01 | 2.3158E+00 | 1.5819E+00 | 1.3111E+00 | 1.3460E+00 | 3.7543E+00 | 1.4289E+00 | 1.5244E+00 |

**Table A2.** Experimental results of MSIGOA and competitors in CEC 2022 (D=20)

| Function | Index | MSGOA | MRFO | FTTA | NMPA | EOSMA | BEESO | FDBARO | IDE-EDA | APSM-JSO |
|---|---|---|---|---|---|---|---|---|---|---|
| | Best | 3.0000E+02 | 1.7322E+03 | 1.3099E+03 | 3.1435E+02 | 5.8999E+02 | 5.4909E+03 | 2.9534E+03 | 3.0358E+02 | 3.0488E+02 |
| F1 | Mean | 3.0003E+02 | 6.0765E+03 | 6.1459E+03 | 4.9426E+02 | 1.5019E+03 | 1.5537E+04 | 9.2418E+03 | 6.0173E+02 | 6.4230E+02 |
| | Std | 1.6602E-01 | 2.9378E+03 | 3.3601E+03 | 2.1133E+02 | 7.3502E+02 | 4.8387E+03 | 3.9133E+03 | 3.3458E+02 | 4.6660E+02 |
| F2 | Best | 4.0002E+02 | 4.0057E+02 | 4.0014E+02 | 4.3521E+02 | 4.3270E+02 | 4.0283E+02 | 4.4527E+02 | 4.0069E+02 | 4.0311E+02 |
| | Mean | 4.2891E+02 | 4.5538E+02 | 4.5467E+02 | 4.5762E+02 | 4.5634E+02 | 4.4603E+02 | 4.6716E+02 | 4.4925E+02 | 4.4878E+02 |
| | Std | 2.4170E+01 | 1.4482E+01 | 1.5736E+01 | 1.1347E+01 | 1.1247E+01 | 1.3489E+01 | 2.3388E+01 | 1.8466E+01 | 1.8829E+01 |
| F3 | Best | 6.0011E+02 | 6.0005E+02 | 6.0070E+02 | 6.0241E+02 | 6.0003E+02 | 6.0004E+02 | 6.0022E+02 | 6.0005E+02 | 6.0004E+02 |
| | Mean | 6.0172E+02 | 6.0926E+02 | 6.0313E+02 | 6.0630E+02 | 6.0020E+02 | 6.0031E+02 | 6.0228E+02 | 6.0151E+02 | 6.0109E+02 |
| | Std | 2.8692E+00 | 7.4421E+00 | 1.6801E+00 | 2.7420E+00 | 2.6777E-01 | 2.6391E-01 | 1.9992E+00 | 1.3217E+00 | 1.2760E+00 |
| F4 | Best | 8.1592E+02 | 8.3184E+02 | 8.2487E+02 | 8.2583E+02 | 8.1328E+02 | 8.2410E+02 | 8.3002E+02 | 8.1498E+02 | 8.1269E+02 |
| | Mean | 8.3192E+02 | 8.6416E+02 | 8.4926E+02 | 8.5732E+02 | 8.2800E+02 | 8.7439E+02 | 8.5807E+02 | 8.2988E+02 | 8.3733E+02 |
| | Std | 9.5267E+00 | 1.6817E+01 | 1.6043E+01 | 1.5695E+01 | 8.0022E+00 | 2.6162E+01 | 1.4057E+01 | 8.5846E+00 | 1.3366E+01 |
| F5 | Best | 9.0727E+02 | 9.2056E+02 | 9.9918E+02 | 9.0792E+02 | 9.0028E+02 | 9.0011E+02 | 9.3512E+02 | 9.0063E+02 | 9.0091E+02 |
| | Mean | 1.0312E+03 | 1.6974E+03 | 1.3748E+03 | 1.1093E+03 | 9.0541E+02 | 9.1654E+02 | 1.2807E+03 | 9.7222E+02 | 9.4378E+02 |
| | Std | 1.1175E+02 | 5.1276E+02 | 2.9823E+02 | 1.9625E+02 | 1.1009E+01 | 4.6346E+01 | 3.2416E+02 | 1.7706E+02 | 6.8020E+01 |
| F6 | Best | 1.8223E+03 | 1.8471E+03 | 1.8751E+03 | 1.8403E+03 | 1.9090E+03 | 2.0121E+03 | 1.8969E+03 | 1.8853E+03 | 1.8506E+03 |
| | Mean | 1.8608E+03 | 5.5637E+03 | 7.5667E+03 | 1.8807E+03 | 5.5163E+03 | 7.7981E+03 | 4.7460E+03 | 2.5306E+03 | 2.7496E+03 |
| | Std | 2.7449E+01 | 4.5425E+03 | 6.2192E+03 | 2.1872E+01 | 4.0238E+03 | 6.5657E+03 | 3.7996E+03 | 1.9743E+03 | 2.2257E+03 |
| F7 | Best | 2.0223E+03 | 2.0260E+03 | 2.0235E+03 | 2.0270E+03 | 2.0215E+03 | 2.0247E+03 | 2.0246E+03 | 2.0201E+03 | 2.0263E+03 |
| | Mean | 2.0310E+03 | 2.0818E+03 | 2.0776E+03 | 2.0463E+03 | 2.0387E+03 | 2.0476E+03 | 2.0577E+03 | 2.0414E+03 | 2.0505E+03 |
| | Std | 6.5572E+00 | 3.7201E+01 | 5.0056E+01 | 1.1409E+01 | 8.3087E+00 | 1.8447E+01 | 3.2058E+01 | 1.7715E+01 | 1.6328E+01 |
| F8 | Best | 2.2146E+03 | 2.2207E+03 | 2.2208E+03 | 2.2244E+03 | 2.2169E+03 | 2.2231E+03 | 2.2205E+03 | 2.2233E+03 | 2.2246E+03 |
| | Mean | 2.2242E+03 | 2.2360E+03 | 2.2531E+03 | 2.2281E+03 | 2.2284E+03 | 2.2349E+03 | 2.2322E+03 | 2.2334E+03 | 2.2413E+03 |
| | Std | 1.7256E+01 | 3.1903E+01 | 4.9126E+01 | 1.8541E+00 | 2.4782E+00 | 1.6308E+01 | 2.9557E+01 | 2.3041E+01 | 3.2936E+01 |
| F9 | Best | 2.4808E+03 | 2.4808E+03 | 2.4808E+03 | 2.4809E+03 | 2.4808E+03 | 2.4808E+03 | 2.4809E+03 | 2.4808E+03 | 2.4808E+03 |
| | Mean | 2.4808E+03 | 2.4808E+03 | 2.4808E+03 | 2.4810E+03 | 2.4808E+03 | 2.4808E+03 | 2.4828E+03 | 2.4808E+03 | 2.4808E+03 |
| | Std | 2.3292E-04 | 5.4460E-02 | 1.8507E-02 | 1.3539E-01 | 3.3276E-02 | 1.6145E-02 | 2.4476E+00 | 1.7716E-04 | 1.7421E-04 |
| F10 | Best | 2.5004E+03 | 2.5005E+03 | 2.4089E+03 | 2.5004E+03 | 2.5003E+03 | 2.5005E+03 | 2.5005E+03 | 2.5005E+03 | 2.5003E+03 |
| | Mean | 2.6315E+03 | 2.8822E+03 | 2.5539E+03 | 2.5227E+03 | 2.5418E+03 | 2.7947E+03 | 2.6145E+03 | 2.7596E+03 | 3.0557E+03 |
| | Std | 2.8835E+02 | 5.0528E+02 | 1.0716E+02 | 5.5878E+01 | 9.3452E+01 | 2.4106E+02 | 1.9740E+02 | 5.2605E+02 | 7.0890E+02 |
| F11 | Best | 2.9000E+03 | 2.6000E+03 | 2.6002E+03 | 2.9365E+03 | 2.9001E+03 | 2.9018E+03 | 2.6268E+03 | 2.9000E+03 | 2.9000E+03 |
| | Mean | 2.9033E+03 | 2.9098E+03 | 2.9108E+03 | 3.0362E+03 | 2.9005E+03 | 2.9098E+03 | 2.9602E+03 | 2.9017E+03 | 2.9006E+03 |
| | Std | 1.4681E+01 | 1.0248E+02 | 7.8516E+01 | 1.0548E+02 | 1.9566E-01 | 7.8970E+00 | 1.2762E+02 | 5.1310E+00 | 2.9297E+00 |
| F12 | Best | 2.9341E+03 | 2.9405E+03 | 2.9416E+03 | 2.9368E+03 | 2.9338E+03 | 2.9340E+03 | 2.9458E+03 | 2.9367E+03 | 2.9384E+03 |
| | Mean | 2.9521E+03 | 3.0156E+03 | 2.9649E+03 | 2.9473E+03 | 2.9454E+03 | 2.9519E+03 | 2.9714E+03 | 2.9609E+03 | 2.9647E+03 |
| | Std | 1.7541E+01 | 6.5366E+01 | 1.8241E+01 | 6.5723E+00 | 6.0031E+00 | 1.2686E+01 | 2.2198E+01 | 1.9360E+01 | 1.7659E+01 |

**Table A3.** Experimental results of MSIGOA and competitors in CEC 2017 (D=50)

| Function | Metric | MSGOA | MRFO | FTTA | NMPA | EOSMA | BEESO | FDBARO | IDE-EDA | APSM-JSO |
|---|---|---|---|---|---|---|---|---|---|---|
| F1 | Best | 9.8782E+05 | 5.8684E+04 | 5.7414E+06 | 4.4880E+08 | 5.2487E+06 | 7.3126E+06 | 3.8523E+08 | 2.3605E+07 | 1.2193E+07 |
| | Mean | 1.6322E+07 | 4.7807E+05 | 5.9912E+07 | 1.9073E+09 | 4.4476E+07 | 2.2389E+07 | 1.4411E+09 | 1.3441E+08 | 1.0584E+08 |
| | Std | 1.0414E+07 | 7.3316E+05 | 6.5073E+07 | 1.1118E+09 | 1.2224E+08 | 1.2278E+07 | 7.7452E+08 | 1.8380E+08 | 8.5119E+07 |
| F2 | Best | 4.8238E+03 | 9.7003E+04 | 1.1227E+05 | 2.9256E+04 | 6.3450E+04 | 1.1047E+05 | 1.1130E+05 | 4.6125E+04 | 3.3499E+04 |
| | Mean | 1.7192E+04 | 1.5330E+05 | 2.0146E+05 | 5.0546E+04 | 1.0573E+05 | 1.6418E+05 | 1.5768E+05 | 7.3126E+04 | 6.6061E+04 |
| | Std | 5.6101E+03 | 2.3879E+04 | 4.6166E+04 | 9.7577E+03 | 1.9735E+04 | 2.0945E+04 | 1.7555E+04 | 1.6886E+04 | 1.6773E+04 |
| F3 | Best | 4.4881E+02 | 4.6786E+02 | 4.6914E+02 | 6.4008E+02 | 5.4022E+02 | 5.5798E+02 | 7.0254E+02 | 5.1636E+02 | 5.1956E+02 |
| | Mean | 5.6558E+02 | 5.9499E+02 | 6.5514E+02 | 8.0088E+02 | 6.2661E+02 | 6.5796E+02 | 8.5852E+02 | 6.5211E+02 | 6.7033E+02 |
| | Std | 5.9874E+01 | 5.6308E+01 | 7.9753E+01 | 1.0638E+02 | 5.3522E+01 | 4.9002E+01 | 9.9210E+01 | 6.5774E+01 | 8.2070E+01 |
| | Best | 6.4261E+02 | 7.0596E+02 | 6.6572E+02 | 7.1606E+02 | 6.1539E+02 | 6.0548E+02 | 6.9910E+02 | 6.1995E+02 | 6.3795E+02 |
| F4 | Mean | 7.2537E+02 | 8.2778E+02 | 7.6384E+02 | 8.2188E+02 | 6.9656E+02 | 6.7189E+02 | 7.7303E+02 | 7.1936E+02 | 7.6145E+02 |
| | Std | 4.3462E+01 | 4.2855E+01 | 4.2653E+01 | 3.9231E+01 | 4.1724E+01 | 4.1340E+01 | 3.9842E+01 | 4.3975E+01 | 6.1338E+01 |
| | Best | 6.0951E+02 | 6.1623E+02 | 6.0867E+02 | 6.2534E+02 | 6.0279E+02 | 6.0339E+02 | 6.1799E+02 | 6.1013E+02 | 6.0774E+02 |
| F5 | Mean | 6.2150E+02 | 6.4032E+02 | 6.1716E+02 | 6.4183E+02 | 6.0641E+02 | 6.1058E+02 | 6.2827E+02 | 6.2349E+02 | 6.2311E+02 |
| | Std | 7.1715E+00 | 1.1684E+01 | 6.3038E+00 | 7.0582E+00 | 2.6604E+00 | 4.6376E+00 | 5.3967E+00 | 6.8822E+00 | 7.8478E+00 |
| | Best | 9.8788E+02 | 1.1121E+03 | 1.0263E+03 | 1.1239E+03 | 9.1933E+02 | 9.4544E+02 | 1.0933E+03 | 1.0571E+03 | 9.8341E+02 |
| F6 | Mean | 1.2028E+03 | 1.4108E+03 | 1.2176E+03 | 1.2086E+03 | 1.0191E+03 | 1.0593E+03 | 1.2668E+03 | 1.2070E+03 | 1.2171E+03 |
| | Std | 9.4438E+01 | 1.9010E+02 | 9.0099E+01 | 4.9199E+01 | 6.0971E+01 | 6.2797E+01 | 1.1464E+02 | 8.9868E+01 | 1.0522E+02 |
| | Best | 9.4152E+02 | 1.0388E+03 | 9.5978E+02 | 1.0521E+03 | 9.2322E+02 | 8.9879E+02 | 9.8794E+02 | 9.3400E+02 | 9.2525E+02 |
| F7 | Mean | 1.0328E+03 | 1.1320E+03 | 1.0688E+03 | 1.1279E+03 | 1.0020E+03 | 9.7889E+02 | 1.0809E+03 | 1.0171E+03 | 1.0542E+03 |
| | Std | 4.4575E+01 | 5.2500E+01 | 3.6761E+01 | 3.5658E+01 | 3.8093E+01 | 4.0440E+01 | 4.1444E+01 | 4.0553E+01 | 5.5775E+01 |
| | Best | 1.8204E+03 | 5.2879E+03 | 4.3019E+03 | 7.2363E+03 | 1.1622E+03 | 1.6524E+03 | 4.2989E+03 | 3.8679E+03 | 3.2200E+03 |
| F8 | Mean | 5.1218E+03 | 1.5905E+04 | 9.2100E+03 | 1.4032E+04 | 2.2951E+03 | 4.2928E+03 | 9.6122E+03 | 1.2054E+04 | 1.0916E+04 |
| | Std | 1.8788E+03 | 5.0276E+03 | 2.6717E+03 | 4.3824E+03 | 9.7988E+02 | 1.8609E+03 | 2.1314E+03 | 4.5145E+03 | 4.3295E+03 |
| | Best | 5.4915E+03 | 6.0240E+03 | 6.8416E+03 | 7.3068E+03 | 7.1707E+03 | 9.4659E+03 | 5.3665E+03 | 7.8183E+03 | 1.0437E+04 |
| F9 | Mean | 7.0224E+03 | 7.4570E+03 | 8.2126E+03 | 8.9618E+03 | 9.8718E+03 | 1.3854E+04 | 7.5190E+03 | 1.2356E+04 | 1.2854E+04 |
| | Std | 6.8887E+02 | 8.3625E+02 | 8.9110E+02 | 8.5354E+02 | 9.8703E+02 | 1.2525E+03 | 6.1986E+02 | 1.6311E+03 | 9.2008E+02 |
| | Best | 1.2076E+03 | 1.2502E+03 | 1.2546E+03 | 1.4473E+03 | 1.3350E+03 | 1.6353E+03 | 1.3944E+03 | 1.2773E+03 | 1.3567E+03 |
| F10 | Mean | 1.3478E+03 | 1.3568E+03 | 1.4302E+03 | 1.5968E+03 | 1.4910E+03 | 2.1005E+03 | 1.7883E+03 | 1.4221E+03 | 1.4898E+03 |
| | Std | 6.8542E+01 | 1.0520E+02 | 1.0168E+02 | 1.1492E+02 | 8.6783E+01 | 3.6925E+02 | 3.2453E+02 | 9.1869E+01 | 9.3655E+01 |
| | Best | 7.7153E+05 | 8.8175E+05 | 3.1849E+06 | 3.1373E+07 | 9.0957E+05 | 2.8001E+06 | 1.0242E+07 | 8.7645E+05 | 8.7503E+05 |
| F11 | Mean | 5.7420E+06 | 3.1004E+06 | 1.8363E+07 | 1.1144E+08 | 1.2773E+07 | 2.1221E+07 | 4.7647E+07 | 6.7061E+06 | 6.7540E+06 |
| | Std | 3.5421E+06 | 1.7374E+06 | 2.2160E+07 | 4.7433E+07 | 9.8258E+06 | 1.3217E+07 | 3.1606E+07 | 3.8684E+06 | 5.6448E+06 |
| | Best | 7.0721E+03 | 1.6369E+03 | 2.1290E+03 | 6.5100E+04 | 4.7564E+03 | 1.4962E+04 | 1.5129E+04 | 4.1050E+03 | 4.9604E+03 |
| F12 | Mean | 2.0228E+04 | 7.0217E+03 | 1.9475E+05 | 2.8738E+05 | 1.2736E+04 | 4.7574E+04 | 5.9504E+04 | 1.1521E+04 | 1.6276E+04 |
| | Std | 1.1372E+04 | 6.7186E+03 | 8.0019E+05 | 2.6925E+05 | 5.5256E+03 | 2.1056E+04 | 5.3378E+04 | 6.7076E+03 | 9.9637E+03 |
| | Best | 1.5061E+03 | 1.0039E+04 | 5.6114E+03 | 1.6210E+03 | 5.6849E+03 | 5.1122E+04 | 2.1865E+04 | 1.7376E+03 | 1.7213E+03 |
| F13 | Mean | 1.6352E+03 | 1.6693E+05 | 3.6165E+05 | 1.6988E+03 | 2.6172E+04 | 2.9635E+05 | 3.3496E+05 | 1.8661E+04 | 1.2094E+04 |
| | Std | 7.6802E+01 | 1.7473E+05 | 4.7318E+05 | 4.2746E+01 | 1.8956E+04 | 2.1738E+05 | 4.1262E+05 | 2.1949E+04 | 1.5926E+04 |
| | Best | 1.9988E+03 | 1.6896E+03 | 1.7490E+03 | 2.8606E+03 | 2.3057E+03 | 4.6620E+03 | 2.0340E+03 | 2.0149E+03 | 2.4099E+03 |
| F14 | Mean | 2.4167E+03 | 8.7510E+03 | 1.0931E+04 | 4.8235E+03 | 1.0515E+04 | 1.4772E+04 | 9.2125E+03 | 6.0452E+03 | 7.7415E+03 |
| | Std | 2.7244E+02 | 5.7133E+03 | 7.5166E+03 | 1.9617E+03 | 5.0512E+03 | 7.0490E+03 | 5.3511E+03 | 4.1088E+03 | 5.7773E+03 |
| | Best | 1.8213E+03 | 2.5482E+03 | 2.4473E+03 | 2.7121E+03 | 2.1928E+03 | 2.4682E+03 | 2.6124E+03 | 2.1696E+03 | 2.0911E+03 |
| F15 | Mean | 2.6273E+03 | 3.3459E+03 | 3.4693E+03 | 3.3296E+03 | 3.0775E+03 | 3.7688E+03 | 3.4422E+03 | 2.9838E+03 | 3.3679E+03 |
| | Std | 3.3626E+02 | 3.8886E+02 | 5.2551E+02 | 3.5302E+02 | 3.6811E+02 | 7.2311E+02 | 3.8502E+02 | 3.8596E+02 | 4.2367E+02 |
| | Best | 2.1550E+03 | 2.5241E+03 | 2.3544E+03 | 2.3715E+03 | 2.1181E+03 | 2.2119E+03 | 2.1295E+03 | 2.2937E+03 | 2.2004E+03 |
| F16 | Mean | 2.5562E+03 | 3.2727E+03 | 3.2001E+03 | 3.0551E+03 | 2.7505E+03 | 3.2494E+03 | 2.9409E+03 | 2.8122E+03 | 3.0085E+03 |
| | Std | 2.1273E+02 | 3.7237E+02 | 3.7634E+02 | 2.3346E+02 | 2.6942E+02 | 4.7101E+02 | 3.1099E+02 | 2.6208E+02 | 3.1944E+02 |
| F17 | Best | 2.3074E+03 | 1.0063E+05 | 1.3692E+05 | 9.8817E+03 | 1.2599E+05 | 6.0080E+05 | 6.2527E+05 | 2.7726E+04 | 3.2963E+04 |
| | Mean | 1.1273E+04 | 1.5317E+06 | 2.2944E+06 | 3.6995E+04 | 5.1349E+05 | 3.6069E+06 | 2.5462E+06 | 1.9590E+05 | 1.3975E+05 |
| | Std | 1.2772E+04 | 1.2346E+06 | 3.5443E+06 | 2.1543E+04 | 5.4887E+05 | 2.5266E+06 | 1.6325E+06 | 1.9776E+05 | 1.1461E+05 |
| | Best | 1.9741E+03 | 2.4070E+03 | 2.0816E+03 | 2.3669E+03 | 2.7478E+03 | 2.5940E+03 | 2.2135E+03 | 2.1359E+03 | 2.2020E+03 |
| F18 | Mean | 2.0738E+03 | 1.6032E+04 | 1.9511E+04 | 7.1596E+03 | 1.7248E+04 | 1.3867E+04 | 1.7292E+04 | 1.2405E+04 | 1.3735E+04 |
| | Std | 4.8488E+01 | 1.0915E+04 | 1.1581E+04 | 5.4550E+03 | 9.6516E+03 | 1.1200E+04 | 1.0780E+04 | 8.1862E+03 | 1.0663E+04 |
| | Best | 2.2414E+03 | 2.5016E+03 | 2.5101E+03 | 2.6193E+03 | 2.3660E+03 | 2.5109E+03 | 2.2459E+03 | 2.3677E+03 | 2.5274E+03 |
| F19 | Mean | 2.6635E+03 | 3.1637E+03 | 3.1838E+03 | 3.0094E+03 | 2.8168E+03 | 3.7693E+03 | 3.0954E+03 | 3.0924E+03 | 3.2075E+03 |
| | Std | 2.1939E+02 | 3.0257E+02 | 2.8420E+02 | 2.2401E+02 | 2.5652E+02 | 3.6504E+02 | 3.4670E+02 | 3.4758E+02 | 3.1726E+02 |
| | Best | 2.4069E+03 | 2.4452E+03 | 2.4217E+03 | 2.4800E+03 | 2.4097E+03 | 2.4150E+03 | 2.4461E+03 | 2.4062E+03 | 2.4490E+03 |
| F20 | Mean | 2.4794E+03 | 2.5507E+03 | 2.5524E+03 | 2.5644E+03 | 2.4710E+03 | 2.4945E+03 | 2.5451E+03 | 2.4995E+03 | 2.5375E+03 |
| | Std | 3.3103E+01 | 5.0308E+01 | 5.3106E+01 | 3.7344E+01 | 3.2100E+01 | 4.7435E+01 | 4.7198E+01 | 4.5134E+01 | 5.7212E+01 |
| | Best | 2.3161E+03 | 2.3021E+03 | 7.8635E+03 | 2.5737E+03 | 2.3205E+03 | 1.1513E+04 | 2.6330E+03 | 2.3840E+03 | 2.4728E+03 |
| F21 | Mean | 7.9808E+03 | 8.9854E+03 | 9.6364E+03 | 7.5381E+03 | 1.0587E+04 | 1.5687E+04 | 8.9914E+03 | 1.1963E+04 | 1.3901E+04 |
| | Std | 2.3682E+03 | 2.3849E+03 | 1.0306E+03 | 4.0181E+03 | 2.6353E+03 | 1.2032E+03 | 1.4674E+03 | 3.7221E+03 | 2.0207E+03 |
| | Best | 2.8329E+03 | 2.9253E+03 | 2.9458E+03 | 2.9265E+03 | 2.8140E+03 | 2.8734E+03 | 2.9471E+03 | 2.8811E+03 | 2.9249E+03 |
| F22 | Mean | 2.8928E+03 | 3.0621E+03 | 3.0666E+03 | 3.0345E+03 | 2.8796E+03 | 2.9617E+03 | 3.0542E+03 | 3.0026E+03 | 3.0549E+03 |
| | Std | 3.9148E+01 | 6.9808E+01 | 6.6530E+01 | 4.7865E+01 | 3.1348E+01 | 5.8017E+01 | 6.2768E+01 | 6.3318E+01 | 7.1338E+01 |
| | Best | 2.9712E+03 | 3.0589E+03 | 3.1020E+03 | 3.0684E+03 | 2.9659E+03 | 3.1106E+03 | 3.1178E+03 | 3.0432E+03 | 3.0604E+03 |
| F23 | Mean | 3.0498E+03 | 3.2692E+03 | 3.2425E+03 | 3.1902E+03 | 3.0370E+03 | 3.2533E+03 | 3.2428E+03 | 3.1573E+03 | 3.1813E+03 |
| | Std | 3.9645E+01 | 9.9929E+01 | 7.8095E+01 | 4.7651E+01 | 3.2221E+01 | 7.0150E+01 | 6.3085E+01 | 6.9634E+01 | 7.7001E+01 |
| | Best | 2.9925E+03 | 3.0563E+03 | 3.0731E+03 | 3.1309E+03 | 3.0804E+03 | 3.0517E+03 | 3.1303E+03 | 3.0605E+03 | 3.0438E+03 |
| F24 | Mean | 3.0865E+03 | 3.1284E+03 | 3.1518E+03 | 3.3088E+03 | 3.1616E+03 | 3.1454E+03 | 3.3552E+03 | 3.1485E+03 | 3.1484E+03 |
| | Std | 4.3597E+01 | 3.5204E+01 | 3.8647E+01 | 1.1942E+02 | 4.2865E+01 | 3.6229E+01 | 9.4946E+01 | 6.0467E+01 | 5.0019E+01 |
| | Best | 2.9469E+03 | 2.9882E+03 | 3.6168E+03 | 3.3049E+03 | 3.3459E+03 | 4.9346E+03 | 3.8678E+03 | 3.7348E+03 | 3.7364E+03 |
| F25 | Mean | 5.5416E+03 | 7.4090E+03 | 7.7951E+03 | 4.4687E+03 | 5.4867E+03 | 6.1937E+03 | 7.3515E+03 | 7.0589E+03 | 7.1098E+03 |
| | Std | 2.4586E+03 | 3.2465E+03 | 1.7407E+03 | 1.0334E+03 | 8.5534E+02 | 5.1095E+02 | 2.0556E+03 | 1.4311E+03 | 8.6249E+02 |
| | Best | 3.2987E+03 | 3.4293E+03 | 3.3455E+03 | 3.4070E+03 | 3.2924E+03 | 3.2917E+03 | 3.3984E+03 | 3.3421E+03 | 3.2811E+03 |
| F26 | Mean | 3.4571E+03 | 3.7073E+03 | 3.5273E+03 | 3.5945E+03 | 3.4108E+03 | 3.4809E+03 | 3.6629E+03 | 3.5316E+03 | 3.5474E+03 |
| | Std | 9.2929E+01 | 1.3252E+02 | 1.0403E+02 | 1.3567E+02 | 7.6028E+01 | 9.1825E+01 | 9.6682E+01 | 1.2557E+02 | 1.1772E+02 |
| | Best | 3.2948E+03 | 3.2978E+03 | 3.3430E+03 | 3.4114E+03 | 3.3464E+03 | 3.3778E+03 | 3.5815E+03 | 3.3485E+03 | 3.3150E+03 |
| F27 | Mean | 3.3512E+03 | 3.4021E+03 | 3.4380E+03 | 3.6512E+03 | 3.4731E+03 | 3.5199E+03 | 3.9036E+03 | 3.4762E+03 | 3.4590E+03 |
| | Std | 3.9352E+01 | 4.2081E+01 | 4.5673E+01 | 1.4606E+02 | 6.9875E+01 | 6.9028E+01 | 1.7431E+02 | 7.5538E+01 | 7.7738E+01 |
| | Best | 3.5381E+03 | 3.7492E+03 | 3.7643E+03 | 3.7945E+03 | 3.7085E+03 | 3.5905E+03 | 3.6997E+03 | 3.7135E+03 | 3.7238E+03 |
| F28 | Mean | 4.0959E+03 | 4.6313E+03 | 4.5885E+03 | 4.5615E+03 | 4.2509E+03 | 4.2666E+03 | 4.4546E+03 | 4.3863E+03 | 4.5623E+03 |
| | Std | 2.6423E+02 | 3.9591E+02 | 3.4499E+02 | 3.6539E+02 | 2.7122E+02 | 3.8624E+02 | 3.1557E+02 | 3.6798E+02 | 3.4568E+02 |
| | Best | 1.0343E+06 | 9.9376E+05 | 6.2441E+05 | 5.7242E+06 | 9.1084E+05 | 8.8058E+05 | 3.1540E+06 | 7.7760E+05 | 9.1409E+05 |
| F29 | Mean | 2.1220E+06 | 1.8954E+06 | 1.0632E+06 | 1.0201E+07 | 2.6580E+06 | 2.8588E+06 | 6.4565E+06 | 1.1492E+06 | 1.9426E+06 |
| | Std | 6.5736E+05 | 7.8806E+05 | 3.2776E+05 | 2.4949E+06 | 1.8116E+06 | 1.2139E+06 | 2.2618E+06 | 2.7287E+05 | 7.7030E+05 |

**Table A4.** Experimental results of MSIGOA and competitors in CEC 2017 (D=100)

| Function | Metric | MSGOA | MRFO | FTTA | NMPA | EOSMA | BEESO | FDBARO | IDE-EDA | APSM-JSO |
|---|---|---|---|---|---|---|---|---|---|---|
| | Best | 9.4908E+08 | 2.5743E+08 | 1.0433E+09 | 1.3744E+10 | 2.1120E+09 | 3.9351E+08 | 2.1073E+10 | 5.7069E+09 | 3.2460E+09 |
| F1 | Mean | 2.1402E+09 | 1.0779E+09 | 2.9756E+09 | 2.5046E+10 | 6.5246E+09 | 1.1745E+09 | 4.0587E+10 | 1.2028E+10 | 1.1621E+10 |
| | Std | 5.6200E+08 | 7.4923E+08 | 1.3125E+09 | 7.6397E+09 | 3.0550E+09 | 3.7654E+08 | 8.3472E+09 | 3.8007E+09 | 4.1957E+09 |
| | Best | 1.1831E+05 | 2.9164E+05 | 4.5617E+05 | 1.8882E+05 | 2.6371E+05 | 3.1770E+05 | 3.0571E+05 | 2.0266E+05 | 2.1937E+05 |
| F2 | Mean | 1.6611E+05 | 3.9151E+05 | 6.8758E+05 | 2.3633E+05 | 3.3801E+05 | 3.8521E+05 | 3.4411E+05 | 2.6364E+05 | 2.7951E+05 |
| | Std | 2.4456E+04 | 5.6350E+04 | 1.4326E+05 | 2.4667E+04 | 3.8092E+04 | 3.8617E+04 | 2.3270E+04 | 3.0717E+04 | 2.9271E+04 |
| | Best | 9.6643E+02 | 9.8175E+02 | 1.1183E+03 | 1.7825E+03 | 1.1948E+03 | 1.0567E+03 | 3.1064E+03 | 1.1931E+03 | 1.2865E+03 |
| F3 | Mean | 1.2329E+03 | 1.3029E+03 | 1.4023E+03 | 3.0214E+03 | 1.5833E+03 | 1.4147E+03 | 4.7700E+03 | 2.0138E+03 | 1.8767E+03 |
| | Std | 1.2723E+02 | 1.9413E+02 | 1.6134E+02 | 5.7311E+02 | 2.4995E+02 | 1.6332E+02 | 1.1143E+03 | 3.5251E+02 | 3.9479E+02 |
| | Best | 1.0309E+03 | 1.1780E+03 | 1.0803E+03 | 1.2908E+03 | 1.0820E+03 | 1.0204E+03 | 1.2887E+03 | 9.6482E+02 | 1.0949E+03 |
| F4 | Mean | 1.2067E+03 | 1.3115E+03 | 1.2662E+03 | 1.4534E+03 | 1.2168E+03 | 1.2265E+03 | 1.3994E+03 | 1.2090E+03 | 1.3227E+03 |
| | Std | 8.8565E+01 | 5.7432E+01 | 7.6161E+01 | 6.9847E+01 | 7.5653E+01 | 1.0675E+02 | 6.6516E+01 | 8.0861E+01 | 1.3312E+02 |
| | Best | 6.3084E+02 | 6.3619E+02 | 6.2569E+02 | 6.5137E+02 | 6.1774E+02 | 6.2636E+02 | 6.4207E+02 | 6.3624E+02 | 6.3255E+02 |
| F5 | Mean | 6.4417E+02 | 6.5846E+02 | 6.3964E+02 | 6.6775E+02 | 6.2728E+02 | 6.3492E+02 | 6.5411E+02 | 6.5215E+02 | 6.5111E+02 |
| | Std | 6.7728E+00 | 6.5211E+00 | 6.5167E+00 | 8.0462E+00 | 5.4079E+00 | 4.2673E+00 | 5.9578E+00 | 8.3254E+00 | 9.6179E+00 |
| | Best | 2.1853E+03 | 2.1290E+03 | 2.1236E+03 | 2.2303E+03 | 1.7611E+03 | 1.4832E+03 | 2.2103E+03 | 2.0414E+03 | 2.0594E+03 |
| F6 | Mean | 2.6037E+03 | 2.8155E+03 | 2.5832E+03 | 2.5510E+03 | 1.9858E+03 | 1.6473E+03 | 2.7909E+03 | 2.5397E+03 | 2.5304E+03 |
| | Std | 1.8695E+02 | 2.9872E+02 | 2.2426E+02 | 1.5537E+02 | 1.2082E+02 | 9.4391E+01 | 2.1784E+02 | 2.3403E+02 | 2.4594E+02 |
| | Best | 1.3644E+03 | 1.3771E+03 | 1.4407E+03 | 1.5065E+03 | 1.3249E+03 | 1.3071E+03 | 1.5461E+03 | 1.3878E+03 | 1.4128E+03 |
| F7 | Mean | 1.5567E+03 | 1.7275E+03 | 1.6369E+03 | 1.7529E+03 | 1.4938E+03 | 1.5173E+03 | 1.7087E+03 | 1.5852E+03 | 1.6340E+03 |
| | Std | 7.6817E+01 | 1.0460E+02 | 1.0358E+02 | 1.0702E+02 | 8.9781E+01 | 8.2617E+01 | 9.0486E+01 | 7.3801E+01 | 1.1171E+02 |
| | Best | 1.6557E+04 | 2.5251E+04 | 2.2876E+04 | 3.9495E+04 | 1.3579E+04 | 1.3808E+04 | 2.3308E+04 | 2.5315E+04 | 3.2558E+04 |
| F8 | Mean | 2.0878E+04 | 3.5021E+04 | 3.0983E+04 | 5.0960E+04 | 2.3968E+04 | 2.5779E+04 | 3.2481E+04 | 5.1117E+04 | 5.2520E+04 |
| | Std | 2.4767E+03 | 1.2272E+04 | 5.2438E+03 | 6.2082E+03 | 5.6940E+03 | 7.0344E+03 | 3.7884E+03 | 9.6236E+03 | 1.0041E+04 |
| | Best | 1.3730E+04 | 1.3124E+04 | 1.5714E+04 | 1.8604E+04 | 2.0122E+04 | 2.9243E+04 | 1.6513E+04 | 2.1989E+04 | 2.5850E+04 |
| F9 | Mean | 1.6036E+04 | 1.6224E+04 | 1.8618E+04 | 2.2608E+04 | 2.3849E+04 | 3.1907E+04 | 1.8915E+04 | 2.8889E+04 | 2.9310E+04 |
| | Std | 1.1118E+03 | 1.9220E+03 | 1.5757E+03 | 1.9122E+03 | 1.5883E+03 | 9.1077E+02 | 1.4182E+03 | 1.9706E+03 | 1.3753E+03 |
| | Best | 3.2279E+03 | 2.8949E+04 | 2.2082E+04 | 1.3371E+04 | 2.2785E+04 | 7.7969E+04 | 3.2548E+04 | 6.8722E+03 | 6.5442E+03 |
| F10 | Mean | 5.3463E+03 | 7.3534E+04 | 5.4841E+04 | 2.2689E+04 | 4.0752E+04 | 1.2494E+05 | 7.1000E+04 | 1.9416E+04 | 1.9277E+04 |
| | Std | 1.1197E+03 | 1.9682E+04 | 2.0989E+04 | 5.0593E+03 | 8.8827E+03 | 2.2725E+04 | 1.7794E+04 | 6.3890E+03 | 7.8558E+03 |
| | Best | 1.0676E+08 | 2.8183E+07 | 1.0106E+08 | 1.0087E+09 | 1.2714E+08 | 1.9799E+08 | 1.2428E+09 | 7.4623E+07 | 1.3567E+08 |
| F11 | Mean | 2.5331E+08 | 1.1412E+08 | 3.2935E+08 | 2.6589E+09 | 3.4603E+08 | 4.1029E+08 | 2.5398E+09 | 4.9076E+08 | 4.2190E+08 |
| | Std | 1.1104E+08 | 1.2010E+08 | 1.4955E+08 | 8.8721E+08 | 1.3072E+08 | 1.3154E+08 | 9.2287E+08 | 2.9983E+08 | 2.1384E+08 |
| | Best | 2.7970E+04 | 3.9553E+03 | 2.1493E+04 | 1.6266E+07 | 2.7405E+04 | 6.7722E+04 | 1.3758E+06 | 2.0326E+04 | 2.3883E+04 |
| F12 | Mean | 1.2496E+05 | 9.7155E+03 | 1.4623E+06 | 7.2457E+07 | 7.9874E+04 | 5.2222E+05 | 7.4059E+06 | 1.6656E+05 | 7.0087E+04 |
| | Std | 1.1395E+05 | 5.9224E+03 | 4.4436E+06 | 6.1797E+07 | 2.6861E+04 | 2.0771E+05 | 4.5973E+06 | 6.0826E+05 | 4.9799E+04 |
| | Best | 2.4780E+03 | 2.1697E+05 | 4.7057E+05 | 3.9877E+04 | 4.0748E+05 | 1.2961E+06 | 9.9625E+05 | 6.0808E+04 | 1.0458E+05 |
| F13 | Mean | 3.3105E+04 | 1.4288E+06 | 2.6734E+06 | 1.8529E+05 | 1.7137E+06 | 5.4233E+06 | 3.8075E+06 | 6.3896E+05 | 4.9148E+05 |
| | Std | 3.5701E+04 | 7.3825E+05 | 1.8272E+06 | 1.0978E+05 | 1.1055E+06 | 2.9735E+06 | 2.0178E+06 | 4.0018E+05 | 2.7463E+05 |
| | Best | 1.7805E+04 | 1.9824E+03 | 4.3939E+03 | 5.6687E+05 | 5.0509E+03 | 2.1809E+04 | 2.9606E+04 | 4.1361E+03 | 4.0211E+03 |
| F14 | Mean | 5.4139E+04 | 5.6485E+03 | 5.2142E+05 | 3.6284E+06 | 1.5003E+04 | 4.7780E+04 | 1.2859E+05 | 9.5109E+03 | 1.3608E+04 |
| | Std | 2.0494E+04 | 7.5104E+03 | 2.4226E+06 | 2.8934E+06 | 6.6330E+03 | 2.3876E+04 | 1.1411E+05 | 4.5916E+03 | 6.0577E+03 |
| | Best | 4.3874E+03 | 4.7970E+03 | 4.5898E+03 | 5.9802E+03 | 4.5267E+03 | 6.0037E+03 | 5.1526E+03 | 4.6720E+03 | 5.0699E+03 |
| F15 | Mean | 5.4211E+03 | 5.8444E+03 | 6.3241E+03 | 7.2268E+03 | 6.4006E+03 | 8.9226E+03 | 6.5931E+03 | 5.9454E+03 | 6.6124E+03 |
| | Std | 6.2249E+02 | 6.8883E+02 | 8.5411E+02 | 7.5544E+02 | 7.5197E+02 | 1.2587E+03 | 7.0441E+02 | 9.2050E+02 | 1.0609E+03 |
| F16 | Best | 3.7043E+03 | 3.8425E+03 | 4.5460E+03 | 4.1517E+03 | 3.8743E+03 | 4.0727E+03 | 4.0437E+03 | 3.7896E+03 | 4.2333E+03 |
| | Mean | 4.6478E+03 | 4.9693E+03 | 5.7626E+03 | 5.2117E+03 | 4.9399E+03 | 6.5881E+03 | 5.0585E+03 | 4.9157E+03 | 5.2235E+03 |
| | Std | 4.1598E+02 | 6.3284E+02 | 6.1580E+02 | 5.6088E+02 | 4.8670E+02 | 1.1224E+03 | 5.3634E+02 | 5.4402E+02 | 6.6421E+02 |
| | Best | 6.3441E+04 | 3.5667E+05 | 7.2810E+05 | 1.2646E+05 | 7.2188E+05 | 2.6413E+06 | 6.5492E+05 | 1.9892E+05 | 2.7359E+05 |
| F17 | Mean | 2.5404E+05 | 2.2010E+06 | 3.4040E+06 | 4.1540E+05 | 2.3008E+06 | 1.0322E+07 | 3.8656E+06 | 1.0286E+06 | 9.8268E+05 |
| | Std | 1.3400E+05 | 1.0350E+06 | 2.3076E+06 | 2.1355E+05 | 1.1677E+06 | 4.9253E+06 | 2.0289E+06 | 6.9060E+05 | 5.5615E+05 |
| | Best | 2.9859E+04 | 2.1327E+03 | 2.8802E+03 | 2.0307E+06 | 3.0085E+03 | 1.5270E+04 | 6.1755E+04 | 2.5808E+03 | 3.1105E+03 |
| F18 | Mean | 1.7665E+05 | 4.8440E+03 | 8.2174E+04 | 7.7019E+06 | 2.3868E+04 | 1.8174E+05 | 3.0731E+05 | 1.2815E+04 | 1.5901E+04 |
| | Std | 1.2385E+05 | 4.0215E+03 | 2.1541E+05 | 5.4504E+06 | 2.4996E+04 | 1.2402E+05 | 2.0678E+05 | 1.4230E+04 | 1.1673E+04 |
| | Best | 3.5525E+03 | 4.1895E+03 | 4.2305E+03 | 4.0949E+03 | 4.1418E+03 | 5.7551E+03 | 4.1527E+03 | 4.7466E+03 | 4.3492E+03 |
| F19 | Mean | 4.5107E+03 | 5.3254E+03 | 5.5806E+03 | 5.2571E+03 | 5.1563E+03 | 7.3345E+03 | 5.0136E+03 | 6.1967E+03 | 6.4657E+03 |
| | Std | 3.8440E+02 | 5.0458E+02 | 6.2287E+02 | 4.9994E+02 | 4.5147E+02 | 3.7490E+02 | 4.5737E+02 | 6.8388E+02 | 6.4736E+02 |
| | Best | 2.8018E+03 | 2.8622E+03 | 2.9159E+03 | 3.0057E+03 | 2.7085E+03 | 2.8674E+03 | 3.0298E+03 | 2.8969E+03 | 2.8954E+03 |
| F20 | Mean | 2.9224E+03 | 3.0939E+03 | 3.1717E+03 | 3.1499E+03 | 2.9268E+03 | 3.0860E+03 | 3.1745E+03 | 3.1184E+03 | 3.1922E+03 |
| | Std | 7.4351E+01 | 1.1572E+02 | 1.2501E+02 | 6.8721E+01 | 9.3975E+01 | 1.1450E+02 | 8.7026E+01 | 9.4456E+01 | 1.2130E+02 |
| | Best | 1.5237E+04 | 1.5939E+04 | 1.8212E+04 | 4.9334E+03 | 2.1519E+04 | 2.7844E+04 | 1.9479E+04 | 2.5172E+04 | 2.7166E+04 |
| F21 | Mean | 1.8855E+04 | 1.9919E+04 | 2.1698E+04 | 2.4585E+04 | 2.6215E+04 | 3.3444E+04 | 2.1712E+04 | 3.1190E+04 | 3.1395E+04 |
| | Std | 1.3500E+03 | 1.7070E+03 | 1.8283E+03 | 5.0226E+03 | 2.0408E+03 | 1.7383E+03 | 1.1988E+03 | 2.0080E+03 | 1.4558E+03 |
| | Best | 3.2035E+03 | 3.4553E+03 | 3.3777E+03 | 3.5865E+03 | 3.2316E+03 | 3.2880E+03 | 3.5678E+03 | 3.4102E+03 | 3.4995E+03 |
| F22 | Mean | 3.2971E+03 | 3.6781E+03 | 3.5418E+03 | 3.8325E+03 | 3.3637E+03 | 3.5616E+03 | 3.7899E+03 | 3.6674E+03 | 3.8308E+03 |
| | Std | 6.2500E+01 | 1.4365E+02 | 8.8722E+01 | 1.0998E+02 | 9.0755E+01 | 1.1248E+02 | 1.0644E+02 | 1.3299E+02 | 1.8186E+02 |
| | Best | 3.7100E+03 | 4.1580E+03 | 4.0103E+03 | 4.2399E+03 | 3.6374E+03 | 3.9459E+03 | 4.3313E+03 | 4.0112E+03 | 4.0620E+03 |
| F23 | Mean | 3.8932E+03 | 4.5992E+03 | 4.2893E+03 | 4.5212E+03 | 3.8248E+03 | 4.2329E+03 | 4.7066E+03 | 4.3527E+03 | 4.5784E+03 |
| | Std | 8.3627E+01 | 2.0784E+02 | 1.5051E+02 | 1.3109E+02 | 8.4837E+01 | 1.2789E+02 | 1.4655E+02 | 1.6928E+02 | 2.2254E+02 |
| | Best | 3.6895E+03 | 3.6457E+03 | 3.7656E+03 | 4.2368E+03 | 3.8979E+03 | 3.8885E+03 | 5.1773E+03 | 3.8737E+03 | 3.9988E+03 |
| F24 | Mean | 3.9199E+03 | 3.8872E+03 | 3.9909E+03 | 5.3217E+03 | 4.3270E+03 | 4.1964E+03 | 6.2163E+03 | 4.5962E+03 | 4.5017E+03 |
| | Std | 1.2080E+02 | 1.2497E+02 | 1.2828E+02 | 4.5124E+02 | 2.1203E+02 | 1.5985E+02 | 5.7997E+02 | 3.3081E+02 | 2.9951E+02 |
| | Best | 5.6800E+03 | 6.6712E+03 | 6.7134E+03 | 8.7066E+03 | 7.9485E+03 | 1.3018E+04 | 1.9833E+04 | 1.4784E+04 | 1.5645E+04 |
| F25 | Mean | 1.5997E+04 | 2.2404E+04 | 1.8299E+04 | 1.7309E+04 | 1.3704E+04 | 1.5164E+04 | 2.3497E+04 | 1.9906E+04 | 1.9288E+04 |
| | Std | 6.6577E+03 | 4.7638E+03 | 3.5089E+03 | 2.0507E+03 | 2.6944E+03 | 1.2811E+03 | 2.3059E+03 | 3.0130E+03 | 2.7543E+03 |
| | Best | 3.4890E+03 | 3.8021E+03 | 3.5459E+03 | 3.6637E+03 | 3.5495E+03 | 3.5423E+03 | 4.0451E+03 | 3.6392E+03 | 3.5854E+03 |
| F26 | Mean | 3.7205E+03 | 4.0984E+03 | 3.7625E+03 | 4.0644E+03 | 3.7404E+03 | 3.6847E+03 | 4.3139E+03 | 3.9316E+03 | 3.9764E+03 |
| | Std | 1.4856E+02 | 1.9155E+02 | 1.2626E+02 | 1.8465E+02 | 9.6857E+01 | 8.1839E+01 | 1.7770E+02 | 1.9076E+02 | 1.6851E+02 |
| | Best | 3.6478E+03 | 3.8387E+03 | 3.9258E+03 | 4.7528E+03 | 4.2161E+03 | 4.3572E+03 | 6.1739E+03 | 4.3339E+03 | 4.3300E+03 |
| F27 | Mean | 3.8802E+03 | 4.2783E+03 | 4.3426E+03 | 6.6475E+03 | 4.8994E+03 | 5.5830E+03 | 8.4318E+03 | 5.6047E+03 | 5.7171E+03 |
| | Std | 9.5335E+01 | 2.7526E+02 | 3.1692E+02 | 1.1368E+03 | 4.3769E+02 | 7.9926E+02 | 1.0356E+03 | 7.6703E+02 | 7.5464E+02 |
| | Best | 6.2654E+03 | 5.8093E+03 | 5.9097E+03 | 7.7080E+03 | 5.8204E+03 | 6.1894E+03 | 6.6247E+03 | 6.8064E+03 | 6.6208E+03 |
| F28 | Mean | 7.3229E+03 | 7.4538E+03 | 7.2685E+03 | 8.8315E+03 | 7.6944E+03 | 7.3418E+03 | 8.1192E+03 | 7.8880E+03 | 8.2242E+03 |
| | Std | 5.7917E+02 | 5.8405E+02 | 5.8946E+02 | 6.2968E+02 | 7.5402E+02 | 6.6525E+02 | 6.5004E+02 | 6.1381E+02 | 7.7122E+02 |
| | Best | 1.1481E+06 | 9.3455E+04 | 1.3738E+05 | 2.9038E+07 | 7.9273E+05 | 1.0690E+06 | 6.6463E+06 | 1.2597E+05 | 3.9574E+05 |
| F29 | Mean | 4.6884E+06 | 5.1211E+05 | 2.1693E+06 | 1.0068E+08 | 2.8822E+06 | 5.0769E+06 | 2.7869E+07 | 9.5283E+05 | 2.1272E+06 |
| | Std | 1.9938E+06 | 4.6979E+05 | 7.9761E+06 | 3.9909E+07 | 1.4678E+06 | 2.9686E+06 | 2.1278E+07 | 5.8537E+05 | 1.6374E+06 |